 \documentclass[sigconf, screen, nonacm]{acmart}
\AtBeginDocument{%
  \providecommand\BibTeX{{%
    \normalfont B\kern-0.5em{\scshape i\kern-0.25em b}\kern-0.8em\TeX}}}

\setcopyright{acmcopyright}
\copyrightyear{2023}
\acmYear{2023}
\acmDOI{XXXXXXX.XXXXXXX}

\usepackage{multirow}
\usepackage[ruled]{algorithm2e} 

\usepackage{subcaption}
\usepackage[scaled=.8]{beramono}
\usepackage{fancyvrb}

\usepackage[normalem]{ulem}

\begin{document}

\title{Are Large Language Models Geospatially Knowledgeable?}
\author{Prabin Bhandari}
\email{pbhanda2@gmu.edu}
\orcid{0009-0006-9034-6372}
\affiliation{%
  \department{Department of Computer Science}
  \institution{George Mason University}
  \streetaddress{4400 University Dr}
  \city{Fairfax}
  \state{Virginia}
  \country{USA}
  \postcode{22030}
}

\author{Antonios Anastasopoulos}
\email{antonis@gmu.edu}
\affiliation{%
  \department{Department of Computer Science}
  \institution{George Mason University}
  \streetaddress{4400 University Dr}
  \city{Fairfax}
  \state{Virginia}
  \country{USA}
  \postcode{22030}
}

\author{Dieter Pfoser}
\email{dpfoser@gmu.edu}
\affiliation{%
  \department{Department of Geography and Geoinformation Science}
  \institution{George Mason University}
  \streetaddress{4400 University Dr}
  \city{Fairfax}
  \state{Virginia}
  \country{USA}
  \postcode{22030}
}


\begin{abstract}

Despite the impressive performance of Large Language Models (LLM) for various natural language processing tasks, little is known about their comprehension of geographic data and related ability to facilitate informed geospatial decision-making. This paper investigates the extent of geospatial knowledge, awareness, and reasoning abilities encoded within such pretrained LLMs. 
With a focus on autoregressive language models, we devise experimental approaches related to (i) probing LLMs for geo-coordinates to assess geospatial knowledge, (ii) using geospatial and non-geospatial prepositions to gauge their geospatial awareness, and (iii) utilizing a multidimensional scaling (MDS) experiment to assess the models' geospatial reasoning capabilities and to determine locations of cities based on prompting. Our results confirm that it does not only take larger, but also more sophisticated LLMs to synthesize geospatial knowledge from textual information. As such, this research contributes to understanding the potential and limitations of LLMs in dealing with geospatial information.
\end{abstract}

\ccsdesc[300]{Computing methodologies~Natural language generation}



\keywords{Large Language Models, Geospatial knowledge, Geospatial awareness, Geospatial Reasoning }



\maketitle
\pagestyle{plain}

\section{Introduction}

The recent proliferation of pretrained large language models (LLMs) like GPT-3~\cite{brown2020language} and their impressive performance on several downstream tasks has led the natural language processing~(NLP) community to consider the implicit knowledge these models may contain in their parameters.
Authors have shown that LLMs 
can function, to an extent, as knowledge bases~\cite{petroni-etal-2019-language}, since they store various types of knowledge, such as common sense, relational, and linguistic aspects in their parameters \citep{safavi-koutra-2021-relational, da2021analyzing, peters-etal-2018-dissecting}.
This paper explores whether and to what extent \textbf{geospatial knowledge} is encoded in LLMs and whether such models have \textbf{geospatial awareness}. Finally, we examine the models' \textbf{geospatial reasoning} potential.
Geospatial knowledge includes the factual understanding of geographic data such as location, distance, and area. 
Geospatial awareness is concerned with the ability to perceive and comprehend geographical information.
Finally, geospatial reasoning is the use of geospatial knowledge and awareness for informed decision making.


Most recent and successful LLMs are built using transformer architectures~\cite{vaswani2017attention} specifically designed for sequence tasks such as language modeling. Although transformers consist of an encoder-decoder structure, LLMs usually only use the decoder part since they focus on autoregressive generation.
This autoregressive nature means that LLMs generate coherent and contextually relevant text. Specifically, they predict the next word (token) in a sequence based on the previous words, thereby capturing the dependencies and contextual nuances of the language. 
The autoregressive strategy calculates a probability distribution over the entire vocabulary and, given a token, it samples from this distribution to generate the next token.
We specifically study the LLaMA~\cite{touvron2023llama}, Open Pre-Trained Transformer (OPT)~\cite{zhang2022opt}, and Alpaca~\cite{alpaca} models.



To evaluate LLMs with respect to their geospatial knowledge, awareness, and reasoning capabilities, we conducted the following experiments.
First, we probe the LLMs for actual geo-coordinates of cities. This should provide us with an idea about their concrete geospatial knowledge.
To assess their geospatial awareness, we evaluate whether geospatial prepositions such as ``near'' translate into smaller distances when used in sentences to generate nearby cities as opposed to a control scenario which simply uses the conjunction ``and''.
Last, to gauge the geospatial reasoning potential of LLMs, we perform a multidimensional scaling(MDS)~\cite{borg2005modern} experiment, in which we compare the predicted layout of cities using real distances to a distance measure derived from LLMs.

Our findings reveal that LLMs are becoming more adept at handling and comprehending geospatial data, as evidenced by their encoded geospatial knowledge and subsequent geospatial awareness while generating texts.
Our results also show the possibility of using LLMs in geospatial reasoning tasks.

\section{Related Work}

There has been comparatively little research when it comes to using natural language processing (NLP) techniques for geospatial data.
While NLP has made great progress in a number of areas, its use and usefulness in ``processing'' geographical data have been underutilized.

Previous research in this area has focused on geographical information encoded in word embeddings~\cite{dassereto2019evaluating, lang2019spatial}.
These studies primarily investigated the degree of isomorphism between word embeddings and the geographical concepts existing in the real world.
Our study differs from these studies in the sense that we do not rely on the word embeddings, but rather on a whole language model which facilitates a thorough investigation of LLMs for geospatial data.

Derungs and Purves \cite{derungs2016mining} explored the spatial relationship for ``near'' using the Microsoft N-grams \cite{wang2010overview} to investigate the geospatial awareness of a statistical $n$-gram language model.
They used expressions of ``A near **'' where A stands for different locations and \texttt{**} refers to the autocomplete suggestions generated by the Microsoft $n$-gram language model.
They have shown what is encoded as near in this $n$-gram as near for different scales of locations and generated nearness maps for various locations. In contrast, our work focuses on the currently popular and much better-performing neural language models.

In closely related work, \citet{lietard-etal-2021-language} assessed the geospatial knowledge of LLMs with tasks like geocoordinate prediction and neighboring country prediction. Our research expands on this work by assessing the model's geospatial awareness and the utilization of LLMs for geospatial reasoning tasks.
While their work used smaller models and required some training, our work uses state-of-the-art LLMs models and requires no training, as it leverages the LLMs' zero-shot inference capabilities.
By incorporating all these aspects, we seek to provide a more comprehensive assessment of the geospatial capabilities of LLMs. 



\section{Methodology}

Our methodology involves three different tasks to assess different aspects of the geospatial capabilities of LLMs.

For the first task, evaluating the geospatial knowledge encoded within LLMs, the objective is to correctly predict the locations and coordinates of cities.
The second task, assessing geospatial awareness, analyzes the expressions generated by LLMs when leveraging geospatial prepositions vs. generic expressions, e.g., ``near vs. ``and'' by comparing their resulting respective distances, i.e., are cities that use ``near'' actually closer than when using ``and''?
%
%
Lastly, to assess the LLMs' usefulness for geospatial reasoning, we devise a problem where the goal is to predict the locations of cities based on the relative distances between cities. We generate two ``constellations'', one which uses the actual distances, compared to another one that uses LLM-derived distances.

For all the above tasks, we make use of auto-regressive language models.
We use the foundational language models OPT~\cite{zhang2022opt} and LLaMA~\cite{touvron2023llama} along with the instruction-tuned model Alpaca~\cite{alpaca} in our experimentation.
Instruction-tuned models are language models that have been fine-tuned to produce outputs that are better preferred by humans.
The Alpaca model is based on LLaMA and has been fine-tuned using self-instruct~\cite{wang2022self}.
We limit our investigation to the above open-sourced models rather than closed models like GPT-3 and chatGPT to ensure transparency and reproducability. 
Closed models are proprietary and, as such, limit our ability to examine the training dataset and understand the training process in detail, making it difficult to draw scientifically sound conclusions or conduct comprehensive analyses.
By using open models, we prioritize openness and enable a more thorough investigation of the models' behavior and underlying training data, aligning with our research objectives.

We use HuggingFace's Transformers~\cite{wolf2019huggingface} to implement the LLMs, which we run on an NVIDIA A100 GPU (80GB).
The total estimated GPU run time is around 200 hours.
We use Stanza~\cite{qi2020stanza} to post-process the generated sentences in our contextualizing geospatial prepositions task and use Scikit-learn~\cite{scikit-learn} to implement multi-dimensional scaling (MDS), which is used in the spatial reasoning task in Section~\ref{sec:reasoning} to determine the locations of cities.
Map-based visualizations were done in Python using the Folium~\cite{folium} library with the Stamen Design map style\footnote{\href{https://stamen.com/}{https://stamen.com/}} under the \href{https://creativecommons.org/licenses/by/3.0/}{CC BY 3.0} license. The map data is from OpenStreetMap\footnote{\href{https://www.openstreetmap.org/}{https://www.openstreetmap.org/}} under the \href{https://opendatacommons.org/licenses/odbl/}{ODbL} License.


\section{Measuring Geospatial Knowledge}

This first experiment simply probes LLMs to determine the coordinates (latitude and longitude) of cities.
This task serves as an indicator of the extent to which LLMs encode geospatial knowledge.

\subsection{Experimental Setup}
\textbf{Prompting}, introduced by \citet{brown2020language}, refers to appending a few sample input-output along with a textual prompt to a pre-trained LLM, which is then expected to provide a relevant completion of this input based on the sample inputs and outputs.
This approach is sometimes referred to as in-context learning.
LLMs can handle a wide range of NLP tasks through prompting, eliminating the need for costly model parameter updates through model fine-tuning. 

In our experiments, we use prompts such as the following:\\
\begin{verbatim}
    The geo-coordinates of Peoria are 40.69 and -89.58.
    The geo-coordinates of Oldham are 53.55 and -2.11.
    The geo-coordinates of Plzen are 49.75 and 13.36.
    The geo-coordinates of Kathmandu are ...   
\end{verbatim}

The first three sentences represent a prompt template that illustrates the type of generation that we expect from the model. Our goal is for the LLM to learn from these first three sentences and to complete the last sentence in the best possible way.

The above example is referred to as \textbf{3-shot} inference since we provided three examples prior to the actual sentence completion task. Providing no example is also referred to as \textbf{zero-shot} inference. 
We use both 3-shot and zero-shot inference for the location prediction of cities.
The cities that we use as examples while prompting are selected randomly from a dataset of 3,527 cities (see below). 
We experiment with different prompt templates, as listed below, 
to assess their effect on the generated results:
\begin{Verbatim}[samepage=true]
    Template 1:
        The geo-coordinates of <city> are ...

    Template 2:
        The latitude and longitude of <city> are ...

\end{Verbatim}

For the case of the instruction-following Alpaca model, we provide the following input.
\begin{Verbatim}[samepage=true]
    Below is an instruction that describes a task,
    paired with an input that provides further context. 
    Write a response that appropriately completes the
    request.
    
    ### Instruction:
    Provide the geo-coordinates of the city given below.

    ### Input:
    {city}

    ### Response:
    
\end{Verbatim}

In the above instruction template, we can simply request the latitude and longitude instead of the geo-coordinates, to match the second template that we used for proppting the other models (simply changing the instruction \Verb+geo-coordinates+ with \Verb+latitude+ and \Verb+longitude+) 

\noindent \textbf{Dataset:} \ \  We use the MaxMind database\footnote{\href{https://www.kaggle.com/datasets/max-mind/world-cities-database}{https://www.kaggle.com/datasets/max-mind/world-cities-database}} for a global list of cities that have a population greater than 100k along with their geographic coordinates.
This results in a list of 3,527 cities whose coordinates we predict using LLMs.
 
\noindent \textbf{Experimental Details:} \ \  We use the 6.7B(illion) and 13B-parameter variant of the OPT model and the 7B and 13B-parameter variant of the LLaMA model. 
For the Alpaca model, we instruction-tune the 7B LLaMA model.
We use beam search with five beams as our decoding strategy.
The beam search algorithm is commonly used in language generation tasks to help determine the output sequence that is most likely to occur by keeping a list of several candidate sequences at each step. It investigates many options while taking contextual dependencies into consideration, enabling the creation of cohesive and contextually relevant text.

\begin{table*}[t]
  \caption{Mean error distances in km for coordinate prediction of cities}
  \label{tab:geo-knowledge}
  \begin{tabular}{ccccrr}
    \toprule
    Model & No. of Params & Prompt template & Prompt length &  Mean error distances (km) & Prediction Rate (\%)\\
    \midrule
     Word2Vec & - & - & - & \textbf{2612}  & -\\
     BERT-L & - & - & - & 3077  & -\\
     GPT-2 & - & - & - & 4498  & -\\
    \midrule
     \multirow{8}{*}{OPT} & \multirow{4}{*}{6.7B} & \multirow{2}{*}{1} & zero-shot & 3085  & 21\\
      &  &  & 3-shot & 7714 & 98 \\
      &  & \multirow{2}{*}{2} & zero-shot & 3009  & 43\\
      &  &  & 3-shot & 7304 & 99 \\
      & \multirow{4}{*}{13B} & \multirow{2}{*}{1} & zero-shot & 2452 & 18  \\
      &  &  & 3-shot  & 6830 & 98  \\
      &  & \multirow{2}{*}{2} & zero-shot & 2780 & 22 \\
      &  &  & 3-shot  & 6052 & 98 \\
      \midrule
     \multirow{8}{*}{LLaMA} & \multirow{4}{*}{7B} & \multirow{2}{*}{1} & zero-shot & 1082 & 84 \\
      &  &  & 3-shot & 1634 & 99 \\
      &  & \multirow{2}{*}{2} & zero-shot & 521 & 10 \\
      &  &  & 3-shot & 1469 & 99 \\
      & \multirow{4}{*}{13B} & \multirow{2}{*}{1} & zero-shot & 864 & 89   \\
      &  &  & 3-shot  & 1069 & 99 \\
      &  & \multirow{2}{*}{2} & zero-shot & \textbf{386} & 31  \\
      &  &  & 3-shot  & 1634 & 99 \\
      \midrule
     \multirow{2}{*}{Alpaca} & \multirow{2}{*}{7B} & 1 & - & 1799 & 76  \\
      &  & 2 & - & 2158 & 99 \\
    \bottomrule
  \end{tabular}
\end{table*}

\subsection{Results and Discussion}

The results of our coordinate prediction task are shown in Table~\ref{tab:geo-knowledge}.
The first three rows correspond to the results reported in \citet{lietard-etal-2021-language}, while all subsequent rows reflect our work.
The ``Prompt template'' column mentions which of the two different prompt templates was used.
. 
The ``Prediction rate'' column refers to the percentage of successful predictions made by the LLM, representing the instances where the model correctly generated another city as the result.
We note that, for some models, the generated text does not match the expected output because of the open-ended nature of text generation by auto-regressive LLMs. A failed generation result example is as follows.
\begin{Verbatim}
    The latitude and longitude of Lobito are:
    Lobito is located in Angola.
\end{Verbatim}

The results reported in previous work by \citet{lietard-etal-2021-language} imply that LLMs have limited encoded geospatial knowledge.
In their work, they compare LLMs with Word2Vec~\cite{mikolov2013efficient}, a neural-based word representation in vector space, and report that Word2Vec performed better than LLMs.
Another interesting finding by~\citet{lietard-etal-2021-language} was that BERT~\cite{devlin2018bert}, a bi-directional LLM, performed better than the (larger and later-released) autoregressive GPT-2~\cite{radford2019language}.

\citet{lietard-etal-2021-language} also suggests that larger LLMs might perform better for tasks related to geographic information. This idea is well supported by our results, with the caveat that not all LLMs perform equally well. 
Despite having the same number of parameters, as shown in Table~\ref{tab:geo-knowledge}, the OPT and LLaMA models produced vastly different results in this coordinate prediction task. 
This demonstrates the effect of the model architecture, design decisions, and the pre-training dataset on the model's performance.
Particularly in the zero-shot setting, \textit{the 13-billion variant of LLaMA showed a remarkable 85\% reduction in prediction error for coordinate prediction compared to the best baseline(Word2Vec).}

Overall, larger variants of LLMs generally do perform better, which supports the scaling law principle~\cite{kaplan2020scaling}.
Our research also shows that different prompts produce different outcomes.
For example, the LLaMA models prompted with ``The geo-coordinates of <city> are...'' perform much better than when prompted with ``The latitude and longitude of <city> are...''.
This leaves room for improvement by employing continuous prompts~\cite{li-liang-2021-prefix, liu2021p, liu2021gpt}, at the expense of interpretability.\footnote{Unlike continuous un-interpetable prompts, which are learned embedding vectors), the prompts we currently use are English sentences that any speaker would find realistic}
As part of future work, using trained continuous vectors as inputs instead of natural language text, we may be able to increase the accuracy and prediction rate for coordinate prediction tasks. 

Finally, our results show that prompting or in-context learning can improve the prediction rate, reaching comparable results to the instruction-tuned Alpaca model.
However, the performance in this case is worse than the zero-shot setting, which is somewhat counterintuitive.
We believe that this discrepancy is due to the fact that LLMs might not have seen relevant geo-coordinate examples for certain cities during their pre-training and will ``refrain'' from generating any coordinate predictions for these cities in a zero-shot setting (resulting in the lower prediction rates observed), but overall achieving a higher accuracy.
On the other hand, in the three-shot prompt setting (or in an instruction-following setup), LLMs are almost ``forced'' to provide a (sometimes inaccurate) prediction.

This shows that we can extract the geospatial knowledge encoded in LLMs more efficiently with proper prompt engineering and exposure to diverse geospatial datasets during LLM pre-training. 

In conclusion, our results show that LLMs are becoming more adept at encoding geospatial knowledge, which is also consistent with the scaling law that larger models typically encode more information.
Furthermore, we see that the zero-shot setting outperforms the few-shot setting in terms of accuracy, partly because the few-shot setting leads to higher prediction rates.
These findings highlight the potential for various parameter-efficient fine-tuning methods like continuous prompts to improve LLMs' geospatial knowledge.

\section{Measuring Geospatial Awareness}
\label{sec:awareness}

Geospatial awareness refers to the perception of space and the use of spatial information during everyday activities. 
This idea also applies to generative language models, i.e., the degree to which LLMs capture geospatial information and how this is evident when generating text.
To assess the geospatial awareness of LLMs, we utilize geospatial prepositions, i.e., prepositions that describe spatial relationships between objects or places in a geographical setting.

\subsection{Experimental Setup}
We want the LLM to generate sentences such as \texttt{"<City-A> is near <City-B>"}, where \texttt{"<City-A> is near"} is passed as context.
Assuming that the model has geospatial awareness, \texttt{<City-B>} should be geographically close to \texttt{<City-A>}.

In our experiments, we contextualize the LLM input with a geospatial preposition and evaluate the output the LLM generates and prompt the model as follows:
\begin{Verbatim}
    Albany is near ...
\end{Verbatim}
In the above prompt, \textit{ Albany} is a city and \textit{near} is a geospatial preposition.

We analyze whether the generation of <City-B> given the context of ``<City A> is near'' is affected by the presence of the preposition ``near'' or not.
In addition to ``near'', we also use the prepositional phrases ``close to'' and ``far from''.

We contrast the results of the above experiments with a control experiment where the geospatial preposition is replaced with the conjunction ``and'' using the following prompt:
\begin{Verbatim}
    Albany and ...
\end{Verbatim}
We prompt the model in both zero-shot and three-shot settings.
Additionally, we also append the state of the city in all our inputs as below:
\begin{Verbatim}
    Albany, New York is near ...
\end{Verbatim}

\noindent \textbf{Dataset:} \ \  We curate a list of 93 cities in the contiguous United States to perform an in-depth analysis of the geospatial awareness of LLMs.
The list balances city size with a coverage of most of the contiguous states.

\noindent \textbf{Experimental Setup:} \ \ Based on the results of the coordinate prediction task, we only use the 13B variant of the LLaMA model for this task.
We use ten different prompts per city, where each prompt is created by randomly selecting a city and its closest city in our list. We generate fifty samples for each prompt.
We use top-$k$ sampling-based decoding with $k$ set to 100 and a temperature of 0.9. 
Top-$k$ sampling limits the token pool while decoding to the $k$ most likely options at each step, while temperature controls the randomness during token selection.
A higher temperature value increases randomness, while a lower temperature value reduces randomness.

\subsection{Results and Discussion}

\begin{figure}[t]
  \centering
  \includegraphics[width=\linewidth]{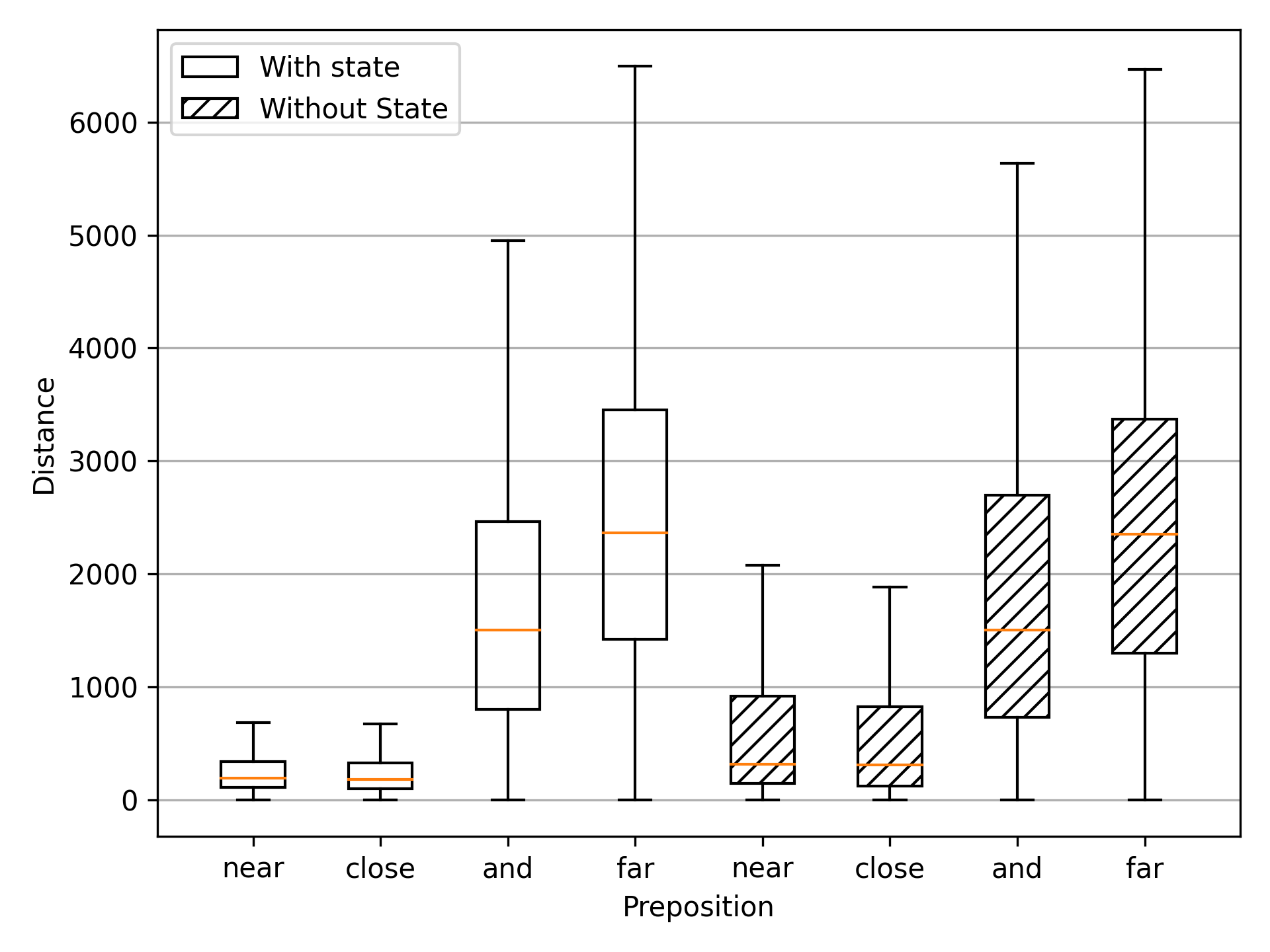}
  \caption{Distances of cities predicted when contextualized with different prepositions.}
   \label{fig:box_plot_prep}
  \Description{Box-plot for the distances of places generated by LLMs when contextualized with different prepositions. The box plots are separated by state and without state.}
\end{figure}

Figure~\ref{fig:box_plot_prep} displays a box plot of the statistics of the actual distances between the generated places and the original city in our experiment. The categories shown are based on the presence or absence of state name in the name of the original city.
The visualization makes it evident that the use of geospatial prepositions in the sentences has an impact on the generated cities.
The sentences contextualized with geospatial prepositions that indicate close proximity, such as ``near" and ``close to", yielded cities that are physically closer to the original city.
Conversely, when the context was ``far'', a geospatial preposition indicating distant location, the generated cities tend to be farther away from the original city.
For our control experiment (with the non-geospatial word ``and''), the observed differences in the distances of the predicted cities provide compelling evidence of the geospatial awareness of LLMs.
The varying magnitude of differences in the distances of predicted cities for the different geospatial prepositions further reinforces the notion of geospatial awareness in LLMs.

\begin{figure*}[t]
    \centering
     \begin{subfigure}[b]{0.435\textwidth}
         \centering
         \includegraphics[trim={10cm 6cm 10cm 5.5cm}, clip,  width=\textwidth]{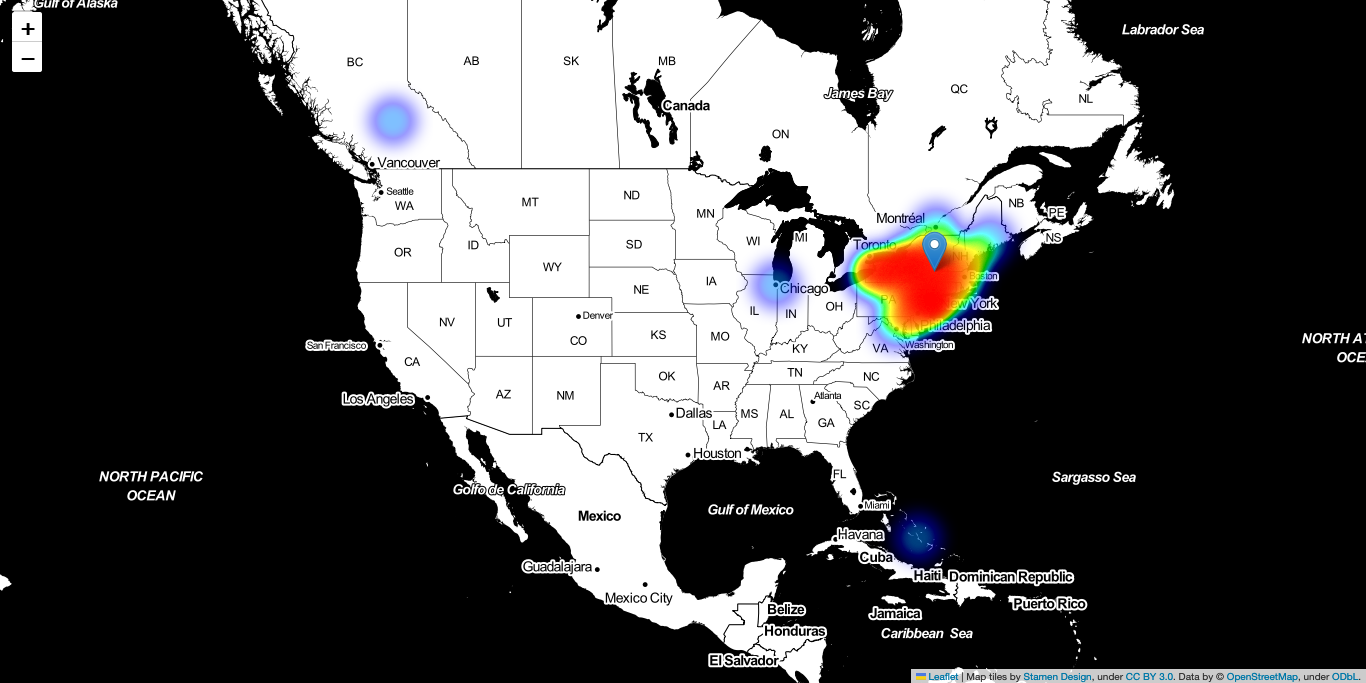}
         \caption{Albany, New York is near ...}
         \label{fig:map_state}
     \end{subfigure}
     \hspace{1cm}
     \begin{subfigure}[b]{0.435\textwidth}
         \centering
         \includegraphics[trim={10cm 6cm 10cm 5.5cm}, clip,  width=\textwidth]{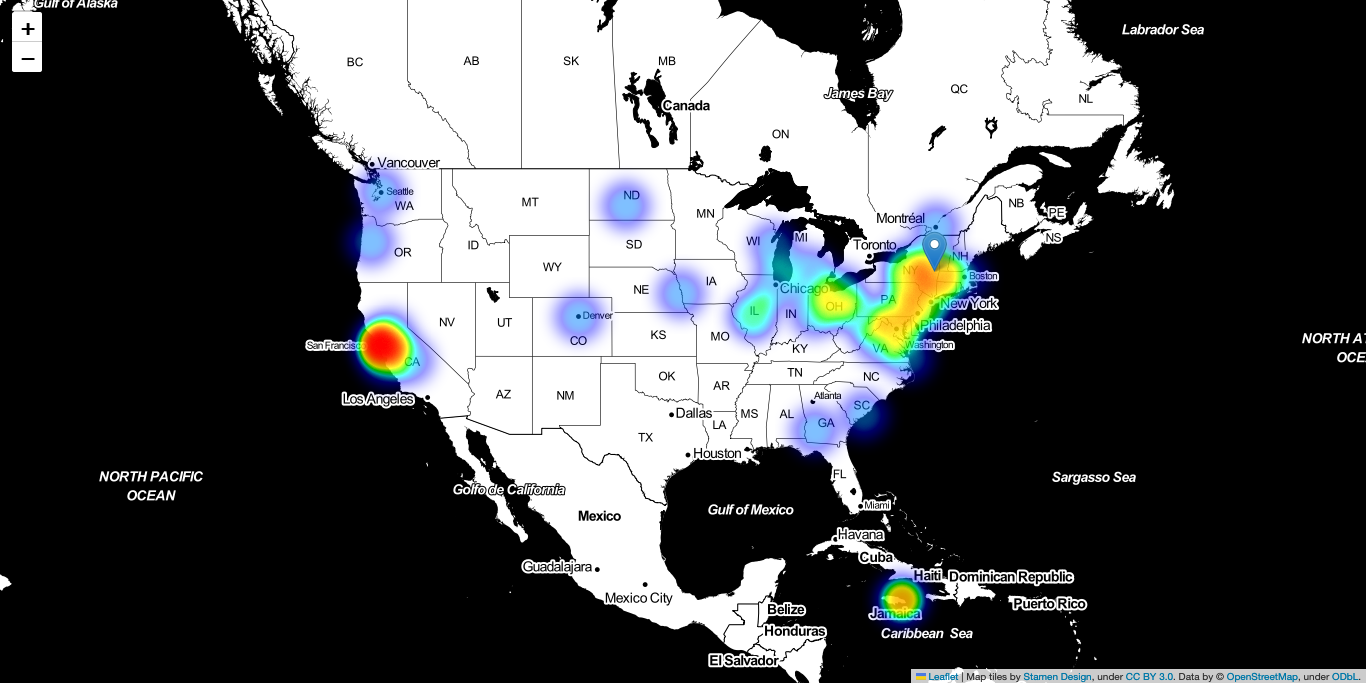}
         \caption{Albany is near ...}
         \label{fig:map_ns}
     \end{subfigure}
    \caption{Heatmaps of the places generated for (a) Albany, New York, and (b) Albany, when contextualized with ``near". When state information is provided, the model can effectively disambiguate between the different similar-named cities.}
  \Description{The figure provides two heatmaps for cities generated by LLM when the context is : (a) Albany, New York is near and (b) Albany is near.}
    \label{fig:heat_map_state}
\end{figure*}

Figure~\ref{fig:heat_map_state} provides a specific example showing that the inclusion or exclusion of the state name in the city names influences the generated cities. 
The generated cities are occasionally further away from the source cities when state names are not included in the prompt.
We believe that this discrepancy is due to the limitation of LLMs in resolving the exact location of a city when the state information is missing: the lack of state information may lead LLMs to confuse cities with the same name (disambiguation).
This situation is demonstrated in Figure~\ref{fig:heat_map_state}, which shows the heatmap of places generated for Albany, New York. 
When the state (NY) is included, the generated cities are almost always located closer to the (correct) Albany in New York State.
In contrast, when the state name is absent from the prompt, the ambiguity between Albany in New York and Albany in California results in two respective hotspots.

\begin{figure*}[t]
    \centering
    \begin{subfigure}[b]{0.495\textwidth}
    \renewcommand\thesubfigure{\alph{subfigure}}
        \centering
        \begin{subfigure}[b]{0.45\textwidth}
         \centering
         \includegraphics[trim={10cm 6cm 10cm 5.5cm}, clip,  width=\textwidth]{plots/maps/albany-near.png}
         \caption{Albany, New York is near}
         \label{fig:map_albany_near}
     \end{subfigure}
     \begin{subfigure}[b]{0.45\textwidth}
         \centering
         \includegraphics[trim={10cm 6cm 10cm 5.5cm}, clip,  width=\textwidth]{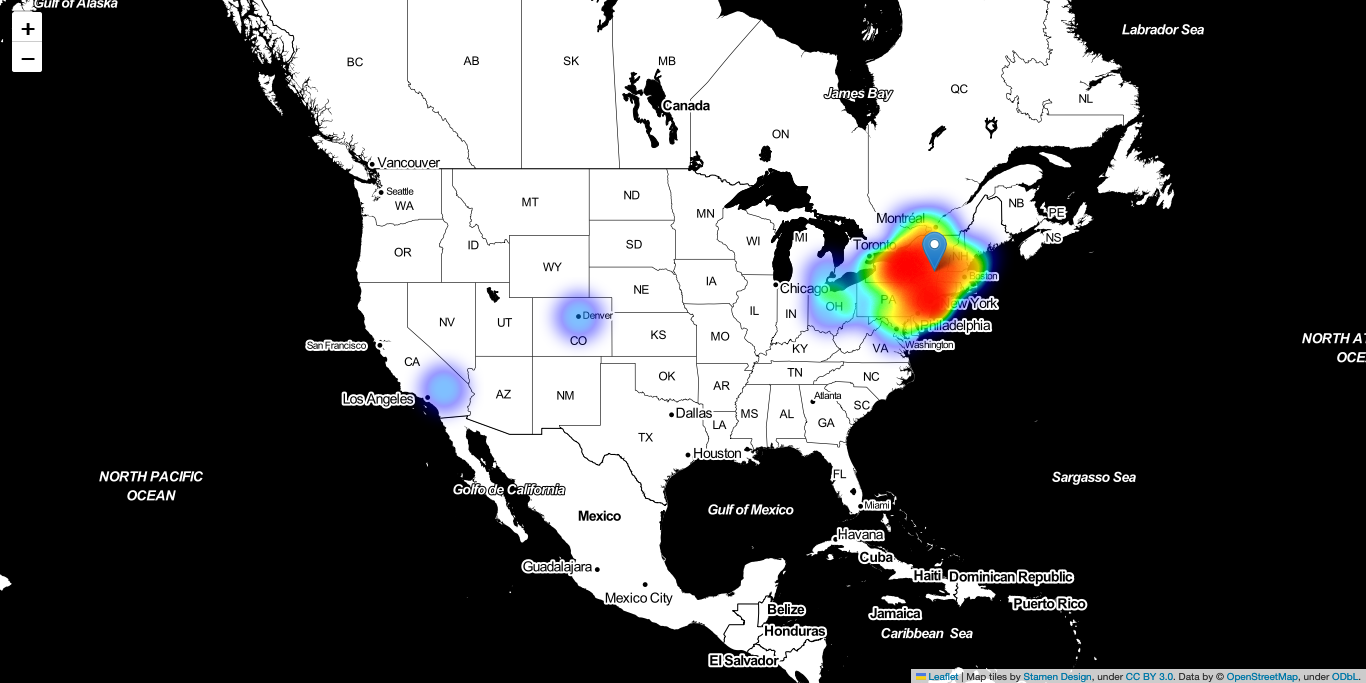}
         \caption{Albany, New York is close to}
         \label{fig:map_albany_close}
     \end{subfigure}
    \begin{subfigure}[b]{0.45\textwidth}
         \centering
         \includegraphics[trim={10cm 6cm 10cm 5.5cm}, clip,  width=\textwidth]{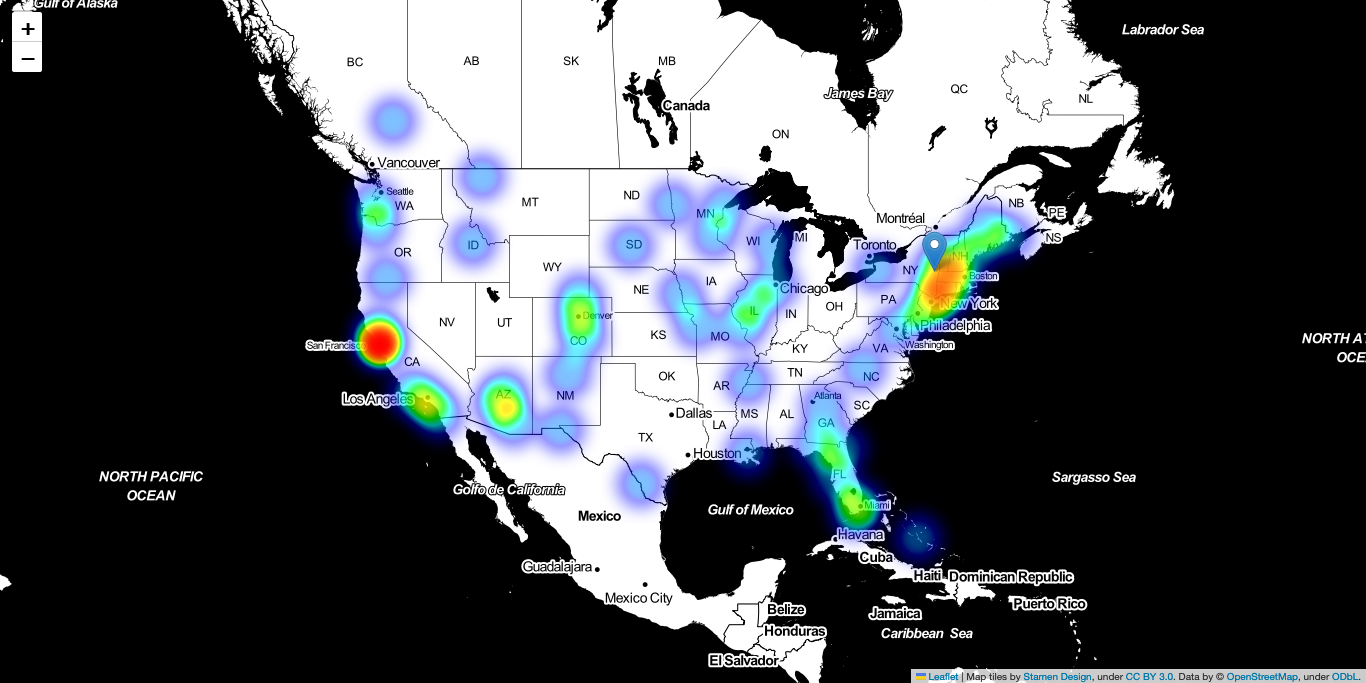}
         \caption{Albany, New York is far from}
         \label{fig:map_albany_far}
     \end{subfigure}
     \begin{subfigure}[b]{0.45\textwidth}
         \centering
         \includegraphics[trim={10cm 6cm 10cm 5.5cm}, clip,  width=\textwidth]{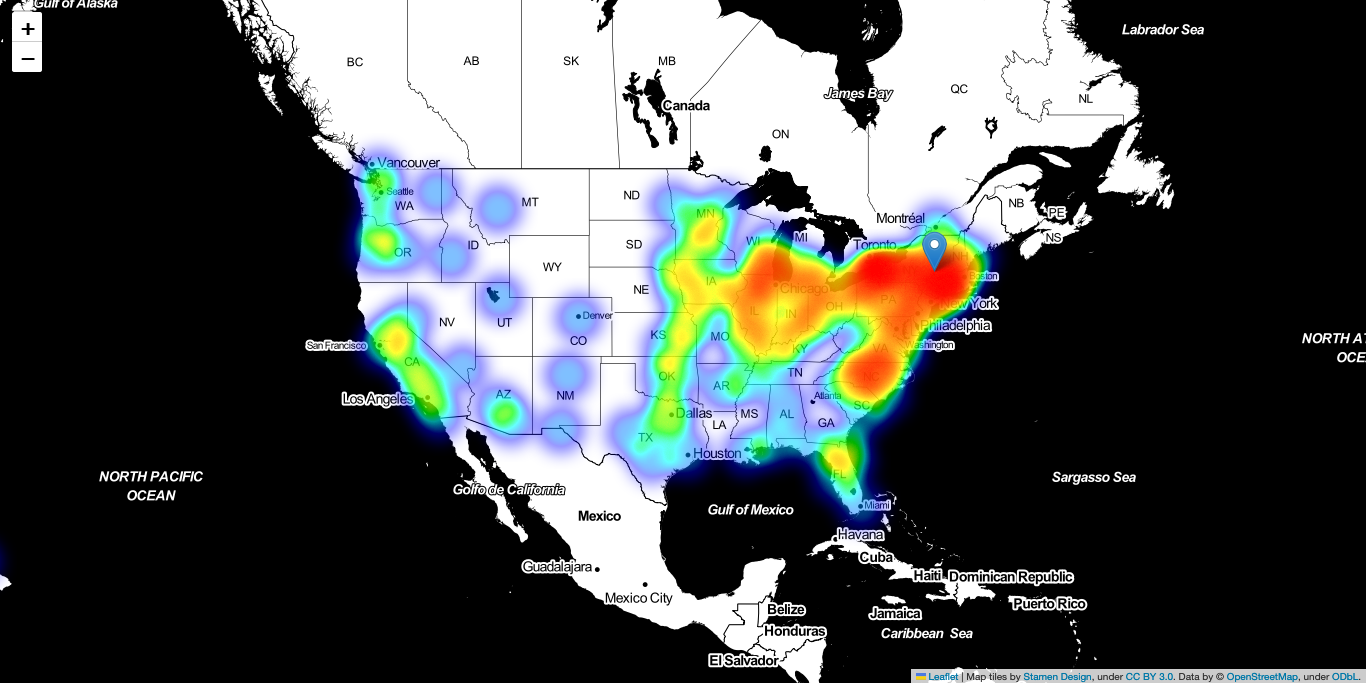}
         \caption{Albany, New York and}
         \label{fig:map_albany_and}
     \end{subfigure}
     \renewcommand\thesubfigure{\roman{subfigure}}
     \setcounter{subfigure}{0}
     \caption{Albany, New York}
     \label{fig:map_a}
    \end{subfigure}
    \begin{subfigure}[b]{0.495\textwidth}
    \setcounter{subfigure}{0}
    \renewcommand\thesubfigure{\alph{subfigure}}
        \centering
        \begin{subfigure}[b]{0.45\textwidth}
         \centering
         \includegraphics[trim={10cm 6cm 10cm 5.5cm}, clip,  width=\textwidth]{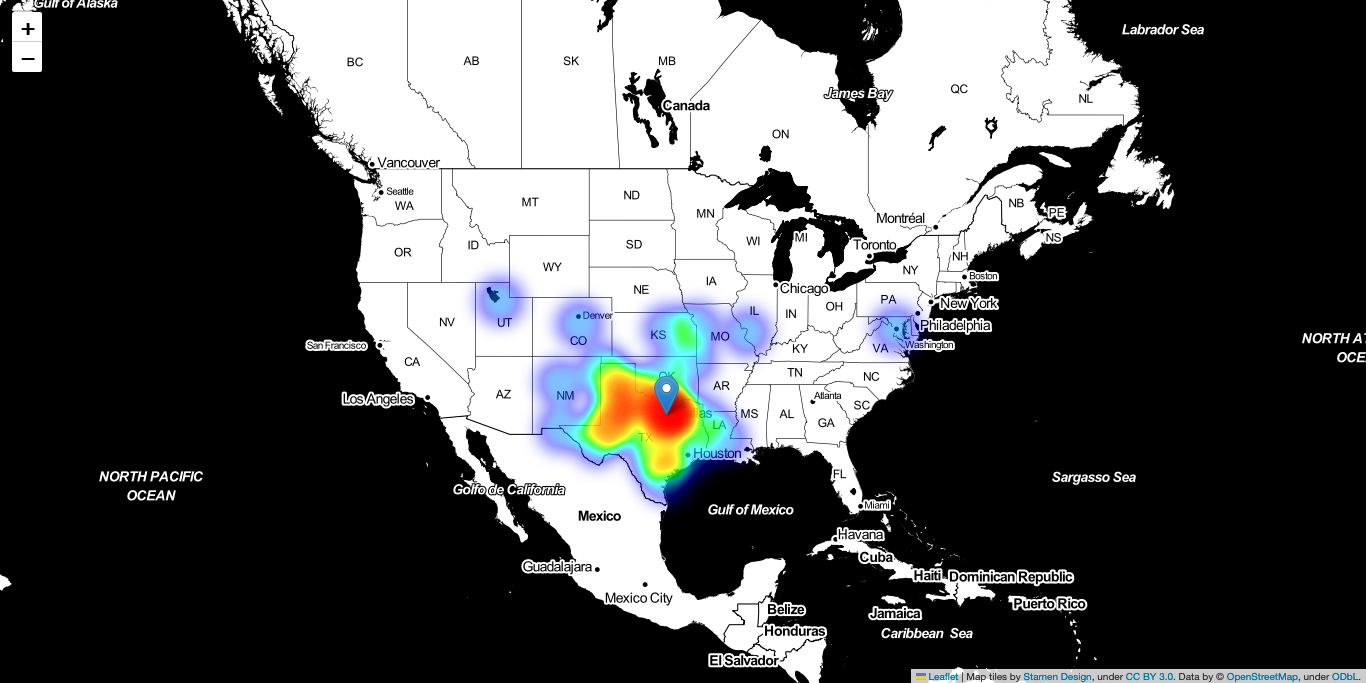}
         \caption{Fort Worth, Texas is near}
         \label{fig:map_fw_near}
     \end{subfigure}
     \begin{subfigure}[b]{0.45\textwidth}
         \centering
         \includegraphics[trim={10cm 6cm 10cm 5.5cm}, clip,  width=\textwidth]{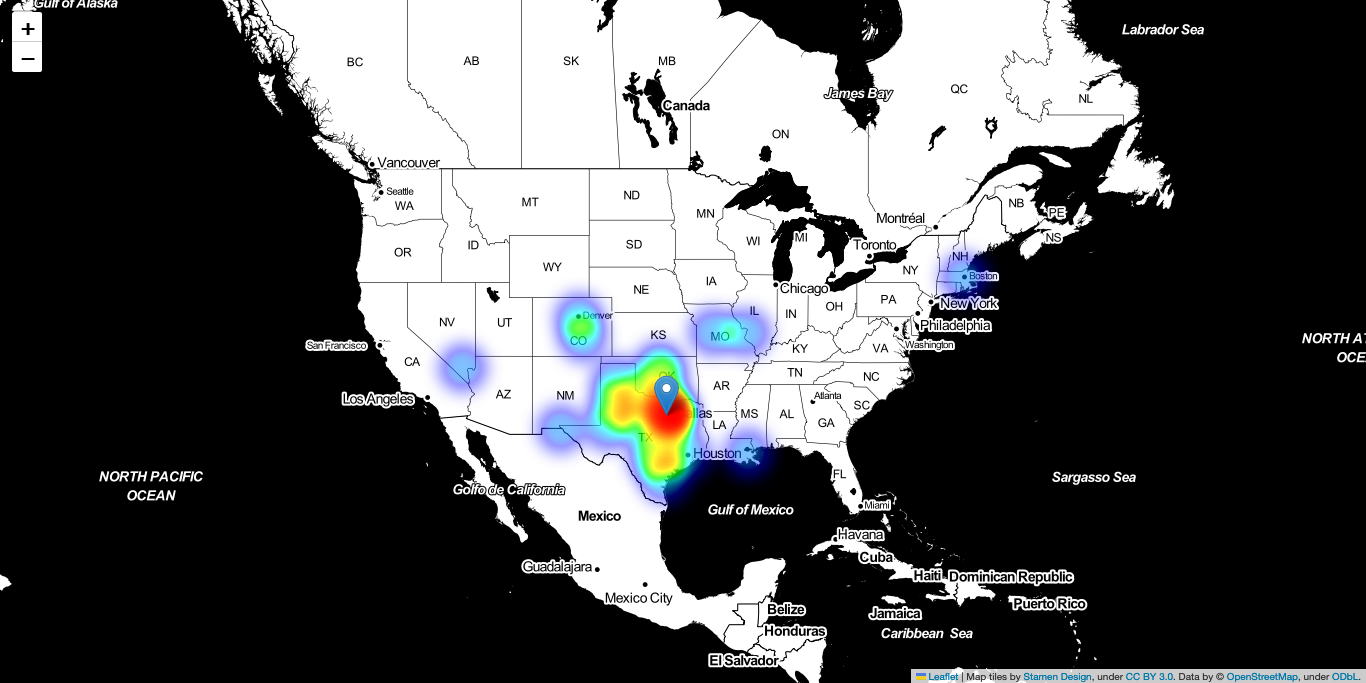}
         \caption{Fort Worth, Texas is close to}
         \label{fig:map_fw_close}
     \end{subfigure}
    \begin{subfigure}[b]{0.45\textwidth}
         \centering
         \includegraphics[trim={10cm 6cm 10cm 5.5cm}, clip,  width=\textwidth]{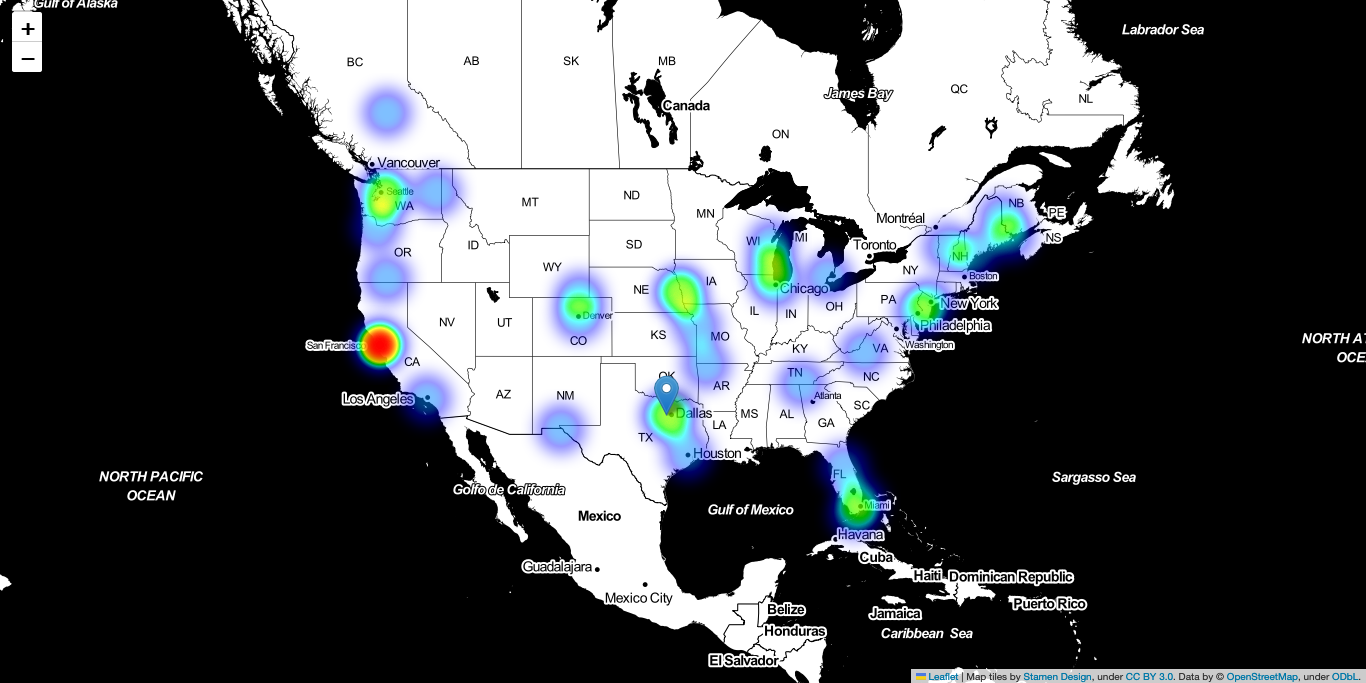}
         \caption{Fort Worth, Texas is far from}
         \label{fig:map_fw_far}
     \end{subfigure}
     \begin{subfigure}[b]{0.45\textwidth}
         \centering
         \includegraphics[trim={10cm 6cm 10cm 5.5cm}, clip,  width=\textwidth]{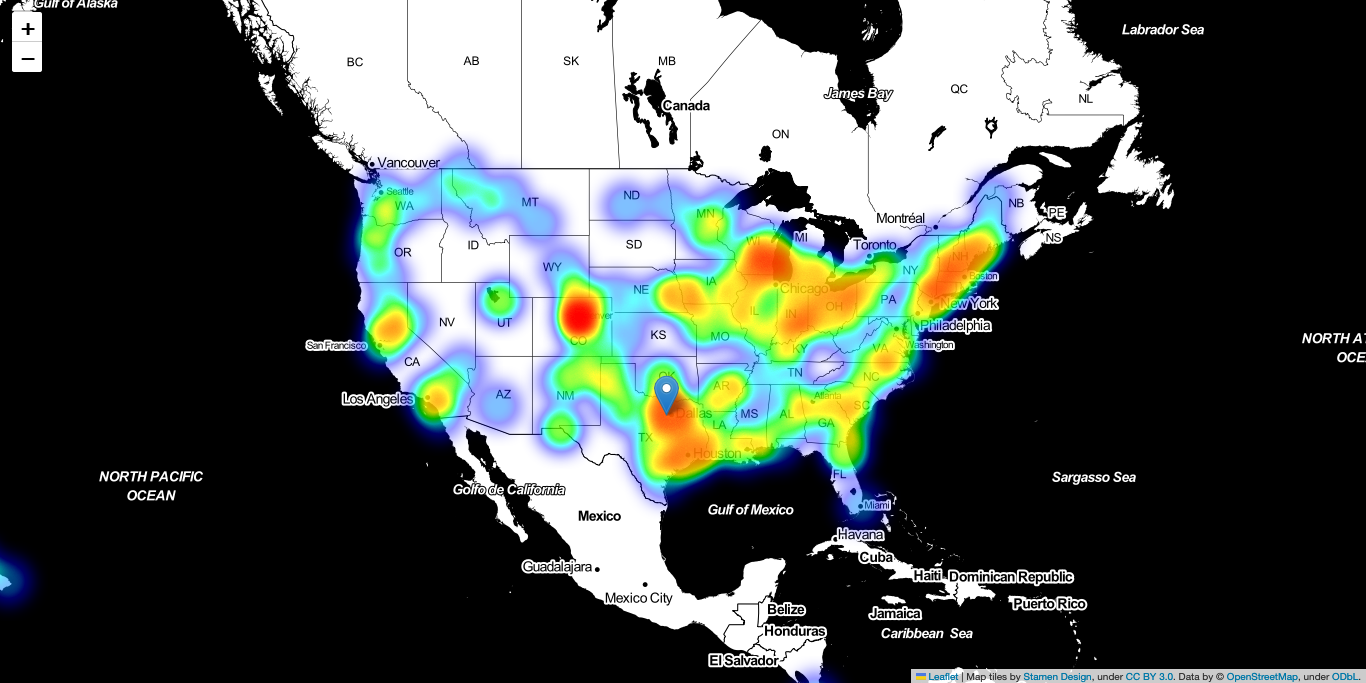}
         \caption{Fort Worth, Texas and}
         \label{fig:map_fw_and}
     \end{subfigure}
     \renewcommand\thesubfigure{\roman{subfigure}}
     \setcounter{subfigure}{1}
     \caption{Fort Worth, Texas}
     \label{fig:map_b}
    \end{subfigure}

    \begin{subfigure}[b]{0.495\textwidth}
    \setcounter{subfigure}{0}
    \renewcommand\thesubfigure{\alph{subfigure}}
        \centering
        \begin{subfigure}[b]{0.45\textwidth}
         \centering
         \includegraphics[trim={10cm 7cm 10cm 4.5cm}, clip,  width=\textwidth]{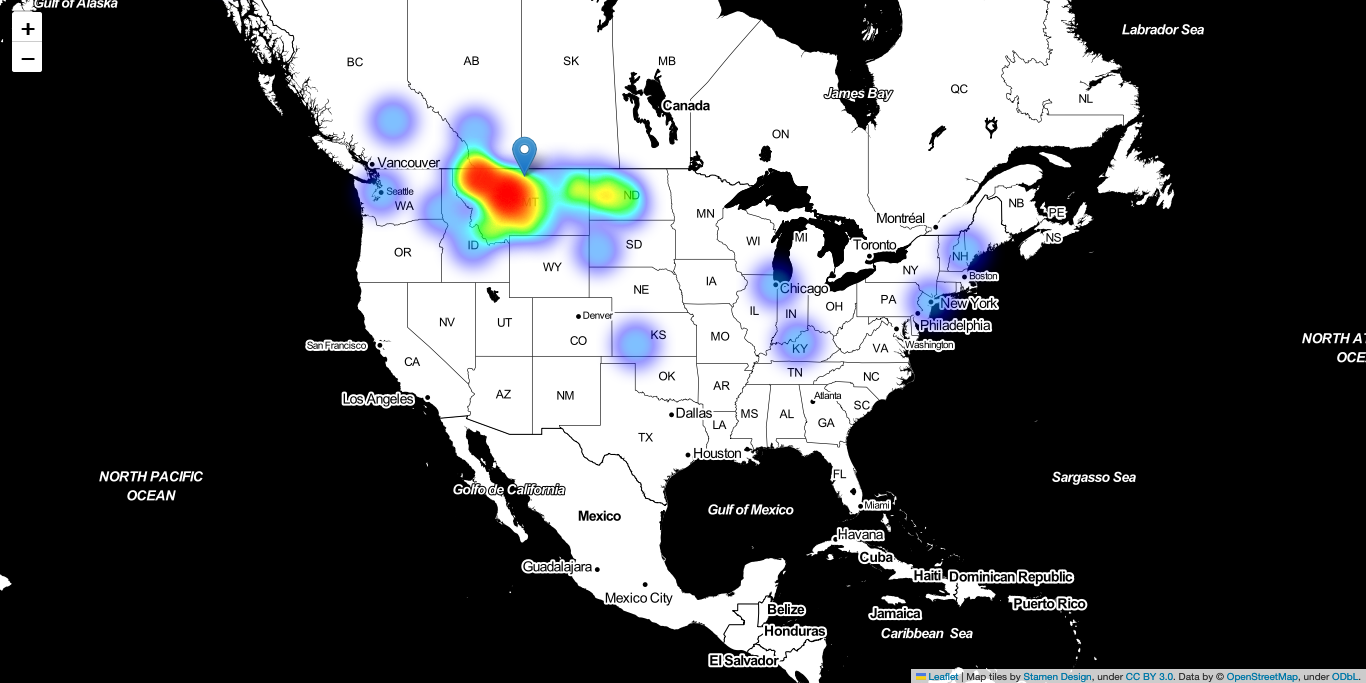}
         \caption{Havre, Montana is near}
         \label{fig:map_havre_near}
     \end{subfigure}
     \begin{subfigure}[b]{0.45\textwidth}
         \centering
         \includegraphics[trim={10cm 7cm 10cm 4.5cm}, clip,  width=\textwidth]{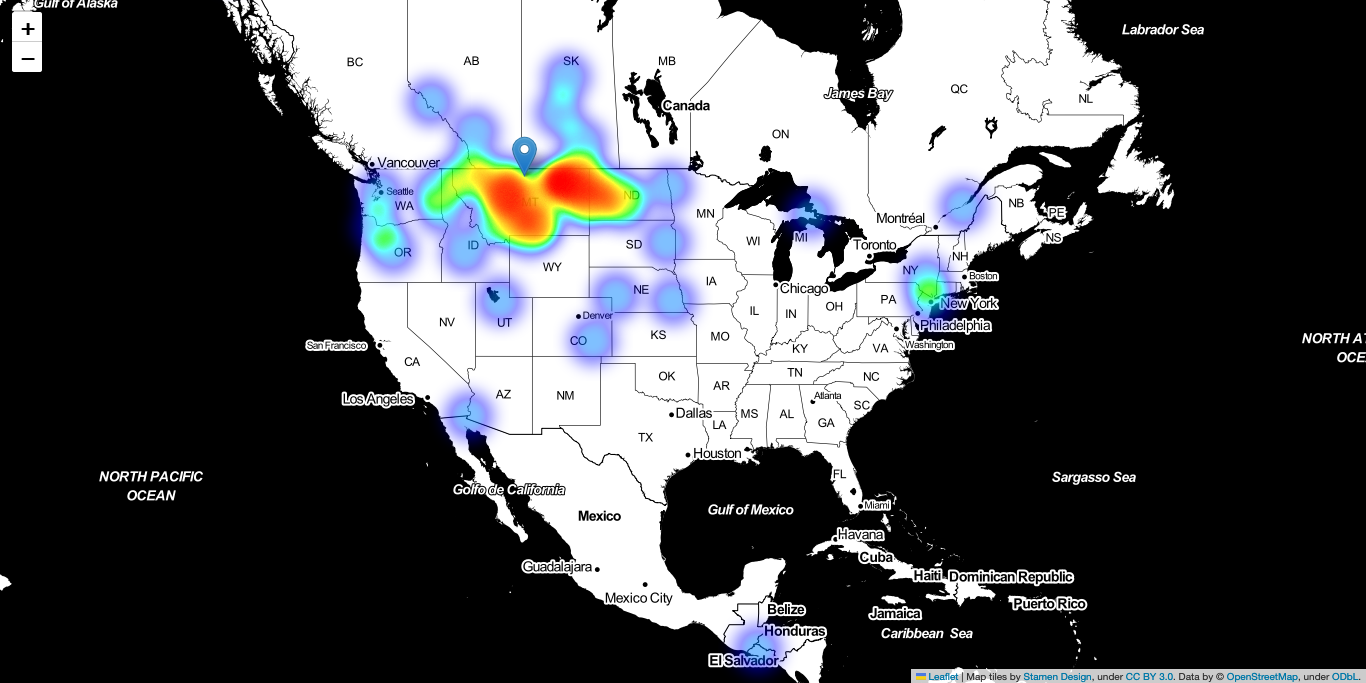}
         \caption{Havre, Montana is close to}
         \label{fig:map_havre_close}
     \end{subfigure}
    \begin{subfigure}[b]{0.45\textwidth}
         \centering
         \includegraphics[trim={10cm 7cm 10cm 4.5cm}, clip,  width=\textwidth]{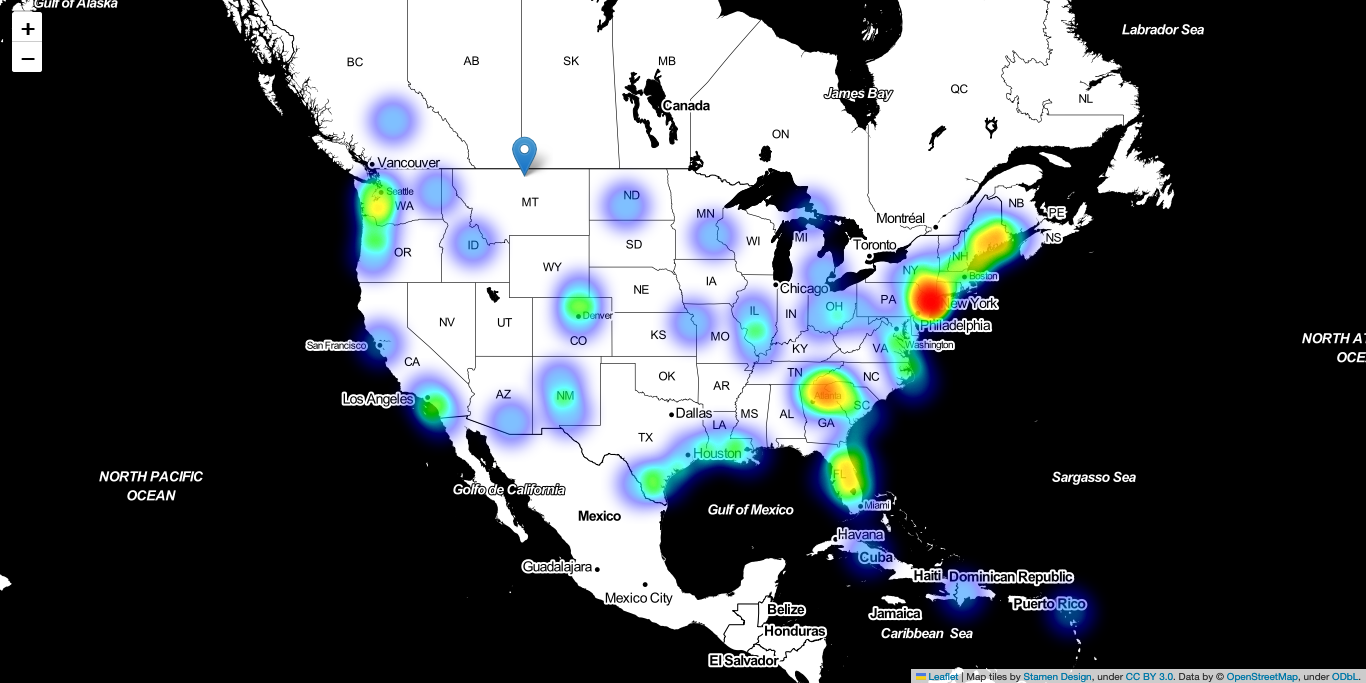}
         \caption{Havre, Montana is far from}
         \label{fig:map_havre_far}
     \end{subfigure}
     \begin{subfigure}[b]{0.45\textwidth}
         \centering
         \includegraphics[trim={10cm 7cm 10cm 4.5cm}, clip,  width=\textwidth]{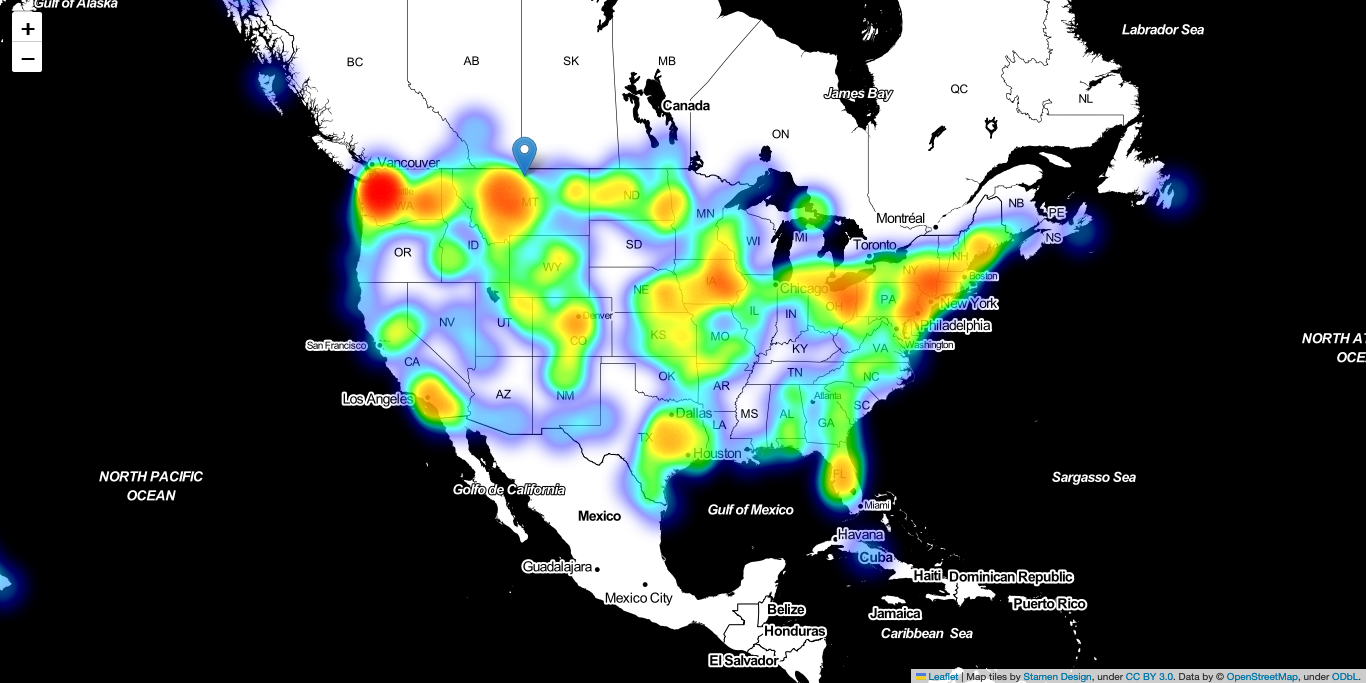}
         \caption{Havre, Montana and}
         \label{fig:map_havre_and}
     \end{subfigure}
     \renewcommand\thesubfigure{\roman{subfigure}}
     \setcounter{subfigure}{2}
     \caption{Havre, Montana}
     \label{fig:map_c}
    \end{subfigure}
    \begin{subfigure}[b]{0.495\textwidth}
    \setcounter{subfigure}{0}
    \renewcommand\thesubfigure{\alph{subfigure}}
        \centering
        \begin{subfigure}[b]{0.45\textwidth}
         \centering
         \includegraphics[trim={10cm 6cm 10cm 5.5cm}, clip,  width=\textwidth]{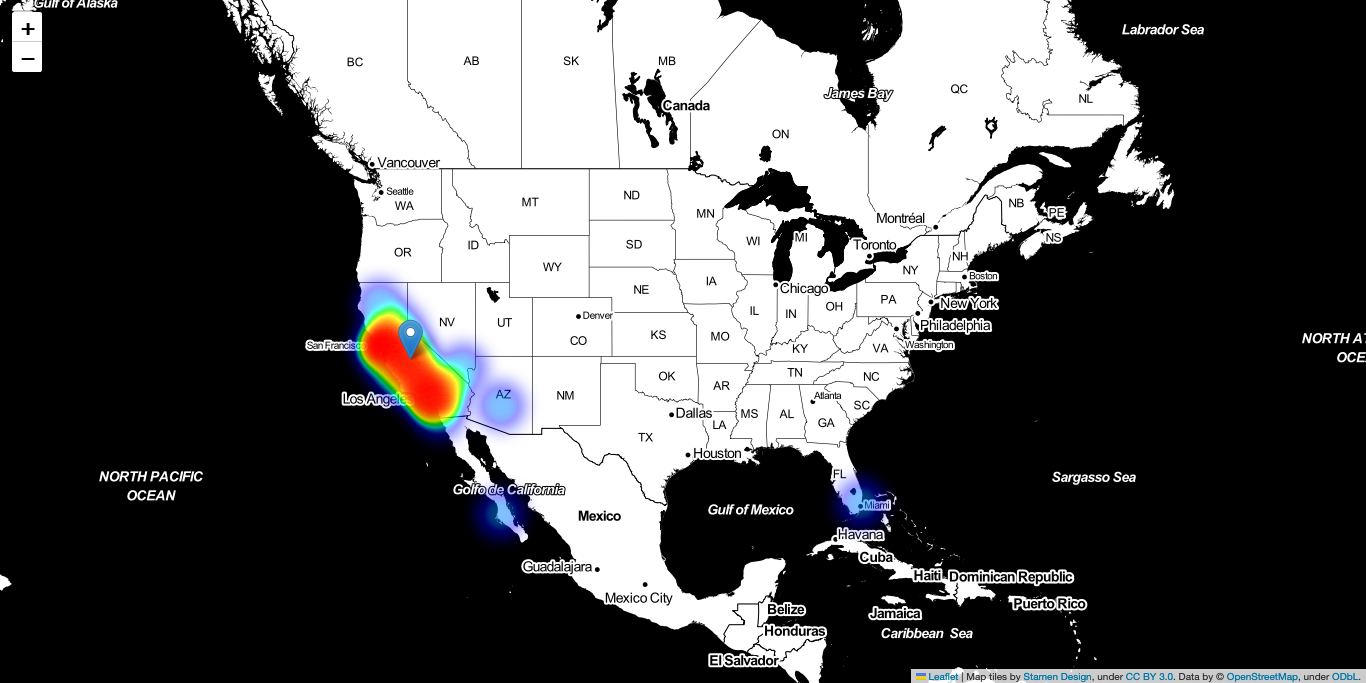}
         \caption{Fresno, California is near}
         \label{fig:map_fresno_near}
     \end{subfigure}
     \begin{subfigure}[b]{0.45\textwidth}
         \centering
         \includegraphics[trim={10cm 6cm 10cm 5.5cm}, clip,  width=\textwidth]{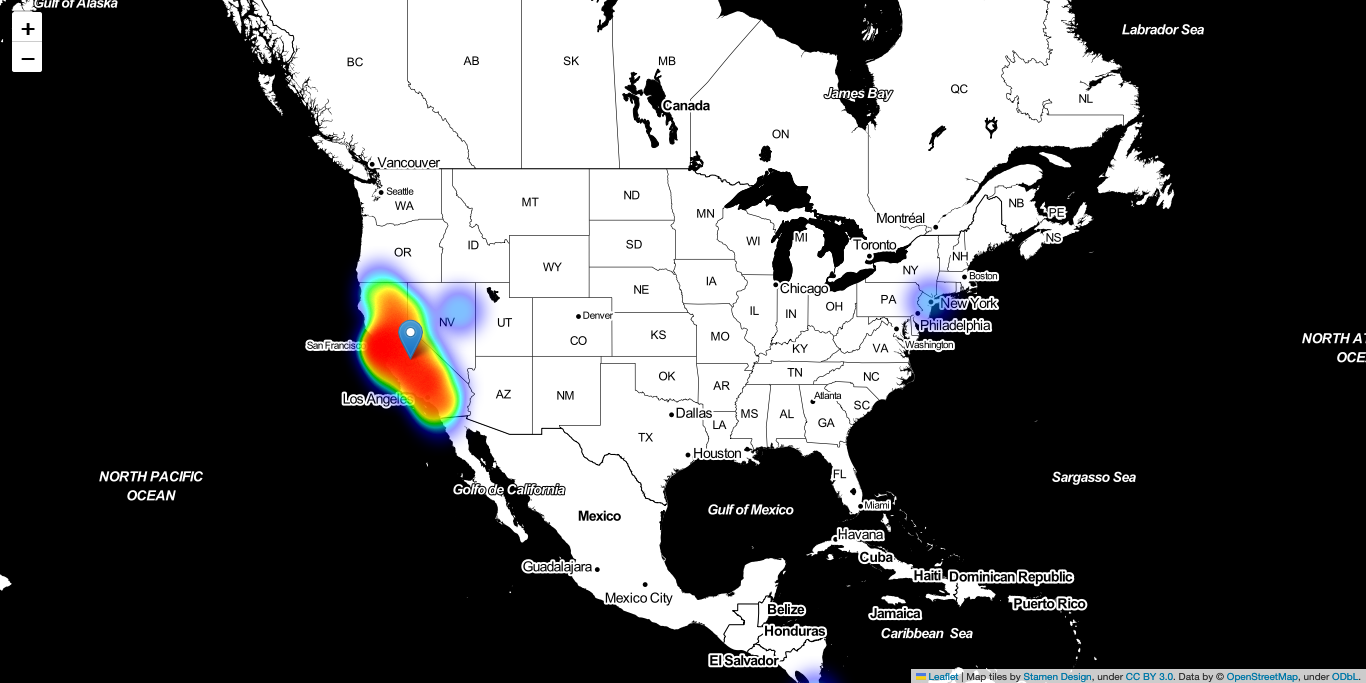}
         \caption{Fresno, California is close to}
         \label{fig:map_fresno_close}
     \end{subfigure}
    \begin{subfigure}[b]{0.45\textwidth}
         \centering
         \includegraphics[trim={10cm 6cm 10cm 5.5cm}, clip,  width=\textwidth]{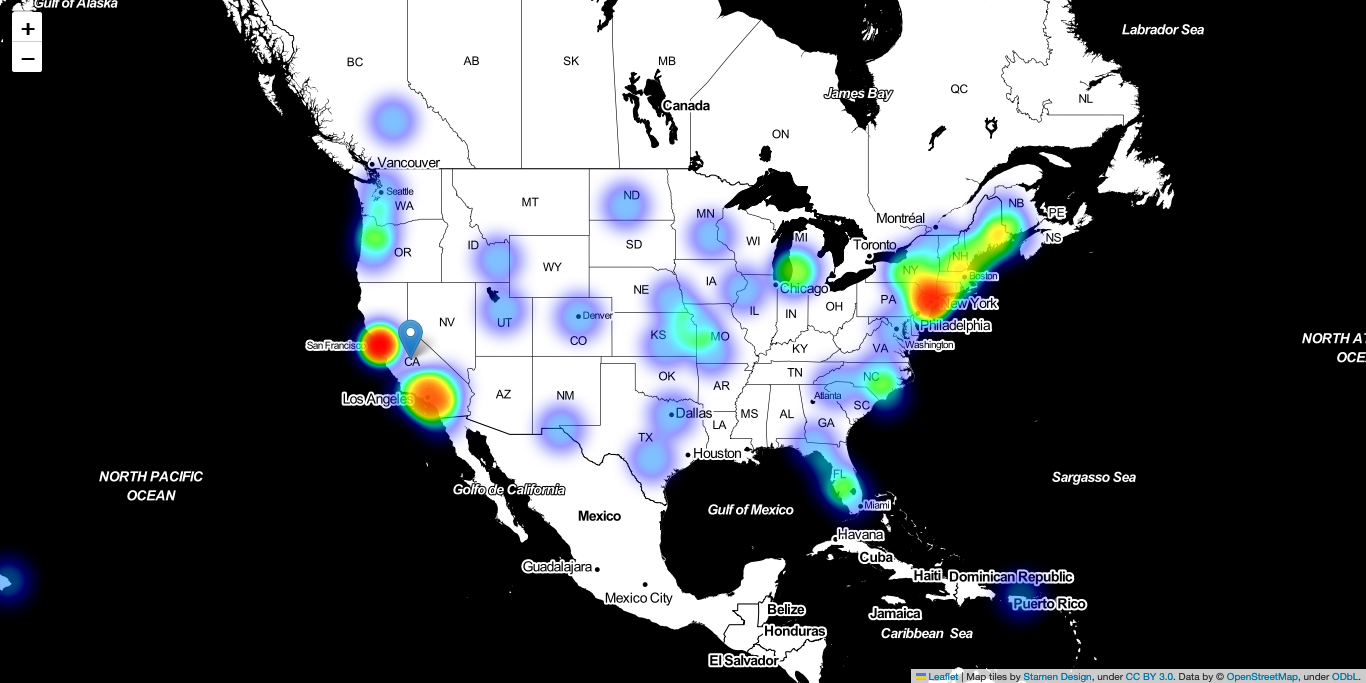}
         \caption{Fresno, California is far from}
         \label{fig:map_fresno_far}
     \end{subfigure}
     \begin{subfigure}[b]{0.45\textwidth}
         \centering
         \includegraphics[trim={10cm 6cm 10cm 5.5cm}, clip,  width=\textwidth]{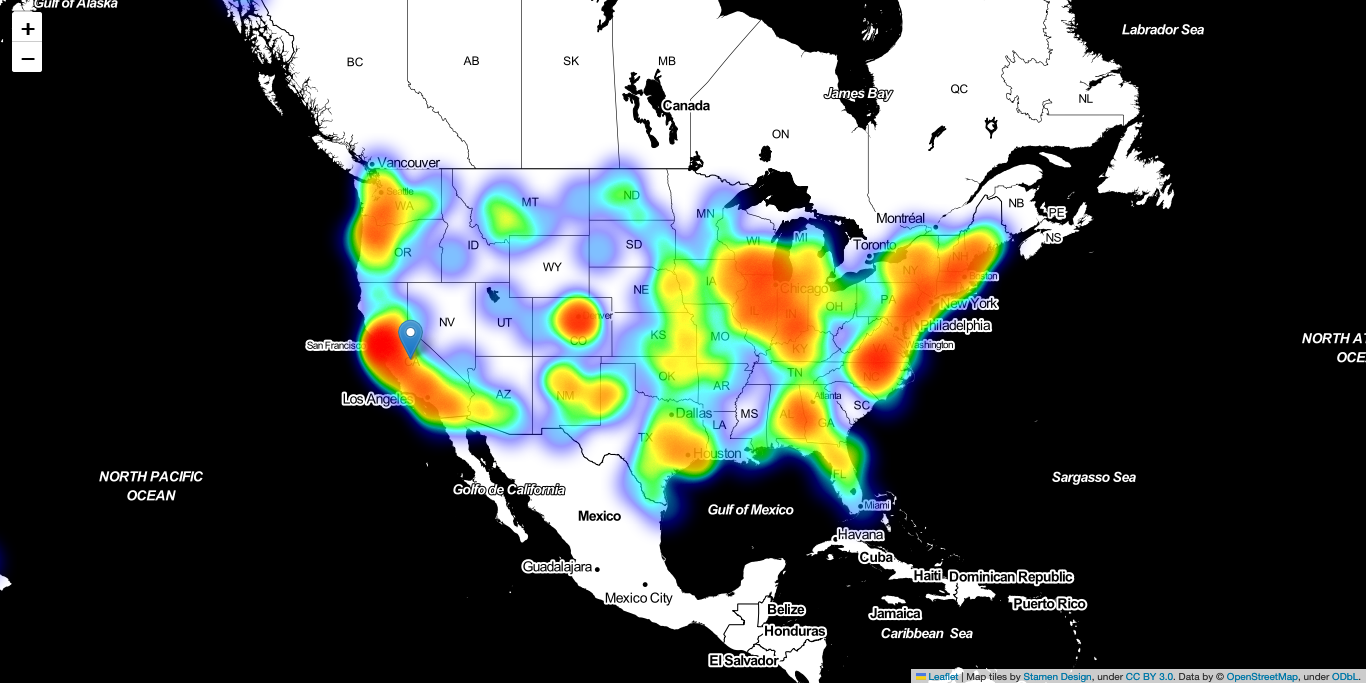}
         \caption{Fresno, California and}
         \label{fig:map_fresno_and}
     \end{subfigure}
     \renewcommand\thesubfigure{\roman{subfigure}}
     \setcounter{subfigure}{3}
     \caption{Fresno, California}
     \label{fig:map_d}
    \end{subfigure}
     
    \caption{Heatmaps of the places generated for (i) Albany, New York, (ii) Fort Worth, Texas, (iii) Havre, Montana, and (iv) Fresno, California when contextualized with various prepositions: (a) ``near", (b) ``close to", (c) ``far from", and (d) ``and".}
  \Description{Different maps showing the heatmaps of the generated cities when the 13B LLaMA model is contextualized with different prepositions for (i) Albany, New York, (ii) Fort Worth, Texas, (iii) Havre, Montana, and (iv) Fresno, California. The  first sub-figure (a) is for "<city> is near". The second sub-figure (b) is for "<city> is close to". The third sub-figure (c) is for "<city> is far from". The fourth sub-figure (d) is for "<city> and".}
    \label{fig:heat_map_cities}
\end{figure*}

Figure~\ref{fig:heat_map_cities} provides additional examples for Albany, New York~(\ref{fig:map_a}), Fort Worth, Texas~(\ref{fig:map_b}), Havre, Montana~(\ref{fig:map_c}), and Fresno, California~(\ref{fig:map_d}). 
In almost all cases, the hotspot is centered on the original city for geospatial prepositions indicating close proximity (``near'', ``close to''), while the hotspots are located further away for prepositions indicating distant places (``far from'').
In contrast, there are no clear hotspots for the control setting (``and'') with predictions spread out across the map, providing further confirmation of the geospatial awareness displayed by LLMs.

\noindent \textbf{The Effect of the Training Data} \ \ 

\citet{elazar2022measuring} explored the effect of training data on LLMs' prediction. They argue that the factual knowledge extracted might be due to the co-occurrence pattern in the dataset that the model was trained on, rather than on a model's hypothesized actual understanding. They suggested that understanding the model's predictions must be accompanied by a study of the training dataset and its effects.

To assess whether the geospatial awareness LLMs exhibit in our experiments
is due co-occurrence patterns in the training dataset, we use the CC100 dataset, which was used for model training, to obtain the occurrence counts\footnote{The co-occurrence count is calculated at the sentence level.} of each city and the co-occurrence counts of all city pairs in our experimentation.

CC100 is a corpus of monolingual data for 100+ languages including English constructed using the methods proposed by \citet{wenzek-etal-2020-ccnet} by processing Commoncrawl~\footnote{\href{https://commoncrawl.org/}{https://commoncrawl.org/}} snapshots. 
Given that 
the LLaMA model was trained directly on Commoncrawl snapshots, we believe that CC100 is a good approximation of the training data.

We conduct an analysis similar to \citet{elazar2022measuring}, by quantifying the number of times (\textit{generation} counts) each city in our list is generated under the different preposition prompts.
We then calculate the Spearman rank correlation coefficient to examine the association between these generation counts with the occurrence of these cities and the co-occurrence frequency of these city pairs in the CC100 dataset.
Additionally, we also consider the correlation of these generations with actual distances, as well as the correlations between distance and co-occurrence to further analyze their relationship. 
All computed rank correlation coefficients are shown in Table~\ref{tab:corr}.

\begin{table}[h]
    \small
  \caption{Spearman rank correlation coefficient between different variables for generations of different prepositions}
  \label{tab:corr}
  \begin{tabular}{@{}c@{ }|r|cccc@{}}
    \toprule
    \multicolumn{2}{c}{} & \multicolumn{4}{c}{Preposition contextualized with}\\
     & Correlation for: &near&close to & far from & and\\
    \midrule
    i & Distance and generation & -0.33 & -0.32 & 0.03 & -0.12 \\
    ii & Co-occurrence and generation & 0.24 & 0.24 & 0.09 & 0.26 \\
    iii & Prompt city and generation & 0.04 & 0.04 & 0.01 & 0.01 \\
    iv & Generated city and generation & 0.13 & 0.14 & 0.13 & 0.30 \\
    v & Distance and co-occurrence & -0.20 & -0.20 & -0.20 & -0.20 \\
  \bottomrule
\end{tabular}
\end{table}

Table~\ref{tab:corr} reveals several interesting findings.
First, there is --as expected-- an inverse correlation between distance and generation (row i) for the geospatial prepositions ``near'' and ``close to'', which is significantly stronger than the correlation obtained for our control word ``and''.
This further solidifies the argument that LLMs possess geospatial awareness, as they generate physically closer places when prompted with geospatial prepositions denoting close proximity.
We did not find any notable correlations for the ``far from'' setting, which can be attributed to the ambiguity associated with this preposition. There is no defined threshold for places to be considered far from each other, in contrast to the close proximity setting. 

Further, as expected we see that the correlation between co-occurrence and generations  (row ii) for ``far from" is low.
However, we observe a positive correlation for the other three prepositions.
We should note that the correlation is slightly stronger for our control word ``and'', compared to geospatial prepositions denoting close proximity.
This shows that, in comparison to prepositions denoting close proximity, co-occurrence between cities has a greater influence on the generations associated with ``and''.
The positive correlation between co-occurrence and generations for ``near'' and ``close to'' can be explained by Tobler's first law of geography~\citep{tobler1970computer}: \textit{``Everything is related to everything else, but near things are more related than distant things''}.
This aligns with the positive correlation observed between distance and co-occurrence.
Therefore, we can confirm that the co-occurrence between cities does not significantly impact the generations for geospatial prepositions any more than what is expected based on Tobler's first law of geography. 
Since the effect is more prominent for our control word ``and'', though we can conclude that the LLMs do possess genuine geospatial awareness.

To further solidify this argument, we also study the correlation between the generation counts and the counts of both, the prompt and generated cities (rows iii and iv).
The little to no correlation between the count of the prompt city and the generations implies that there is a minimal impact of the count of the prompt city on generations.
We do observe a positive correlation between generation and the count of the generated cities, which is expected when sampling from any well-trained LLM.
However, the correlation is considerably more prominent for the control word ``and'' compared to the other geospatial prepositions, which suggests that the frequency of the generated cities in the pre-training dataset plays a more prominent role for the control word ``and'' compared to the geospatial prepositions.
This observation further confirms the claim that LLMs are genuinely geospatially aware.

In conclusion, our results provide compelling evidence that LLMs are indeed geospatially aware. 
Our analysis of the pre-training dataset reinforces the claim by demonstrating that the observed geospatial awareness in LLMs is not merely a result of the patterns seen in the pre-training dataset.

\section{LLMs and Geospatial Reasoning}
\label{sec:reasoning}

Geospatial reasoning refers to the process of understanding and analyzing geospatial information to draw conclusions and make decisions.
In order to assess the usefulness of LLMs for this task, we devise an experiment to predict the locations of cities using dissimilarity measures, such as for example distances between the cities. 

\subsection{Experimental Setup}

We use dissimilarity measures to establish a 2-dimensional geometric representation of cities.
We accomplish this through the application of multi-dimensional scaling (MDS)~\cite{borg2005modern}. 
Specifically, we begin with a list of cities with known locations and with a test city whose location and coordinates we want to predict.
Knowing the distance between all cities (including the test city), 
we then use a least-squares estimation of transformation parameters between two point patterns~\cite{umeyama1991least} to get the transformation matrix that maps the 2-dimensional geometric space coordinates generated by MDS to actual geo-coordinates using the cities for which the geo-coordinates are known.
Finally, we use this transformation matrix to determine the geo-coordinates for the test city.
We provide the detailed pseudocode for this geo-coordinate prediction from a dissimilarity measure task in Algorithm~\ref{alg:geo_pred}.

\begin{algorithm}[h]
\caption{Geo-coordinate prediction from a dissimilarity measure~(distance)}
\label{alg:geo_pred}
\SetKwProg{generate}{Function \emph{predict\_geo\_coordinates(cities, test\_city)}}{}{end}

\generate{}{
    dissimilarity\_matrix = calculate\_dissimilarity\_matrix(cities, test\_city)\;
     mds = MDS(
        number\_of\_components = 2,
        dissimilarity = \textquotesingle precomputed\textquotesingle
     )\;
     known\_coordinates = extract\_geo\_coordinates(cities)\;
     a = mean(known\_coordinates)\;
     init\_value = concatenate(known\_coordinates, a)\;
     x = mds.fit\_transform(dissimilarity\_matrix, init=init\_value)\;
     cities\_2d = x[0:-1] \;
     rotation, translation, scaling = transform (cities\_2d, known\_coordinates) \;
     predicted\_coods = x - mean(x) \;
     predicted\_coods = scaling * predicted\_coods\;
     predicted\_coods = dot\_product(rotation, predicted\_coods.Transpose)\;
     predicted\_coods = predicted\_coods.Transpose + translation\;
     \textbf{return} predicted\_coods\;
     
}
\end{algorithm}

We use actual distances as a benchmark for dissimilarity measures and the co-occurrence counts between each city pair as our baseline measure to establish a comparative reference point.

Co-occurrence is a measure of similarity, so to convert it into a dissimilarity measure, we consider the reciprocal of co-occurrence values.
We adopt a similar approach for the other similarity measure we study, namely, the generation frequency, to convert it into a dissimilarity measure.

By utilizing a dissimilarity measure between cities to predict their geo-location, our designed task illustrates a practical application of geospatial reasoning.
We extract diverse measures of dissimilarity from the LLM and conduct a comparative analysis against our predefined benchmark and baseline. The \textit{dissimilarity measures} include the following:

\begin{itemize}
  \item \textbf{Predicted Distance}: We predict the distances between each city pair in a zero-shot setting from LLM, prompted in a manner similar to the geo-coordinate prediction task (\S4).
  To predict the distances between cities, we prompt the LLM as follows.
  \begin{verbatim}
    The distance in kilometers between Albany,
    New York and Dallas, Texas is ...
\end{verbatim}
  \item \textbf{Generation Frequency}: We count the generation frequency of each city in relation to the remaining cities from our contextualizing geospatial preposition task (\S5).
  The frequencies associated with geospatial prepositions \textit{``near''} and \textit{``close to''} serve as a measure of similarity, which we convert to a dissimilarity measure by taking their reciprocal values.
  Additionally, the frequency corresponding to \textit{``far''} can be considered as a dissimilarity measure.
\end{itemize}

\noindent \textbf{Dataset:} \ \ We use the curated list of 93 cities in the contiguous United States presented in Section~\ref{sec:awareness}.
Each city in our dataset is considered a test city for which we want to predict its coordinates and we use the remaining cities to sample cities with known locations.
Based on the results of Section~\ref{sec:awareness}, we include the state names in the prompts.
 
\noindent \textbf{Experimental Details:} \ \  Similar to our contextualizing geospatial preposition task,  we use the 13B variant of the LLaMA model.
We use beam search with 5 beams as our decoding strategy.

\subsection{Results and Discussion}

We present the result of our location prediction task in Table~\ref{tab:geo-reasoning}.
We report two mean error distances. The first one is based on predicting the geo-coordinate of a city using all the remaining cities in our list from the contiguous US. The second one is calculated by dividing the cities into nine different US census bureau-designated divisions and limiting our experiment to each division both for the known and the unknown cities.

To establish a benchmark, we compute the minimum attainable error, which is obtained by using an ``oracle'', i.e., we use the actual distances for all cities. An additional baseline is the ``random'' one, where we report the average error from ten different random predictions for the test city locations. The latter should function as the maximum ``reasonable'' error.

Our results indicate that limiting the task to a smaller geographical region by using the divisions instead of considering the whole contiguous US leads to better prediction accuracies in general. This can be attributed to the inherent ease of prediction when using cities that are in closer proximity.
Furthermore, the errors associated with co-occurrence counts (row i in Table~\ref{tab:geo-reasoning}) and ``and'' generation counts (row ii) resemble the errors obtained from random distances, which is expected since both of these counts do not convey a similarity or dissimilarity based on proximity between two cities.
The same observation holds --and is even more prominent-- for the ``far from'' generation counts (row v).
On the other hand, the ``near'' and ``close to'' generation counts (rows iii and iv) have much lower errors compared to the co-occurrence count and ``and'' based generations. This further strengthens the argument that LLMs possess geospatial awareness and reasoning capability.
However, due to the sparsity of the generation counts, we do not obtain values that closely align with the predicted distances.
Last, directly asking the LLM to predict distances (row vi) yields results that are much closer to the actual distances and far better than the random distance.

\begin{table}[t]
  \caption{Mean error distances in kilometers for geo-coordinate predictions of cities from dissimilarity measure using MDS}
  \label{tab:geo-reasoning}
  \begin{tabular}{clrr}
    \toprule
    && \multicolumn{2}{c}{Mean error distance (km)} \\
    &Measure & Contiguous & Divisions \\
    \midrule
    \multicolumn{4}{l}{\ \ \ Baselines} \\
    &Actual distance & \textbf{190.41} & \textbf{56.78}\\
    &Random distance & 1440.01 & 483.15\\
    \midrule
    \multicolumn{4}{l}{\ \ \ Our Approach} \\
    i&Co-occurrence count & 1237.01 & 425.64\\
    ii&``and" generations count & 1359.51 & 453.73\\
    iii&``near" generations count & 750.66 & 328.91\\
    iv&``close to" generations count & 782.22 & 324.06\\
    v&``far from" generations count & 1383.23 & 455.76\\
    vi&    Predicted distance & \textbf{346.65} & \textbf{177.41}\\
    \bottomrule
  \end{tabular}
\end{table}

\begin{figure*}[t]
        \centering
        \begin{subfigure}[b]{0.435\textwidth}
         \centering
         \includegraphics[trim={10cm 6cm 10cm 5.5cm}, clip,  width=\textwidth]{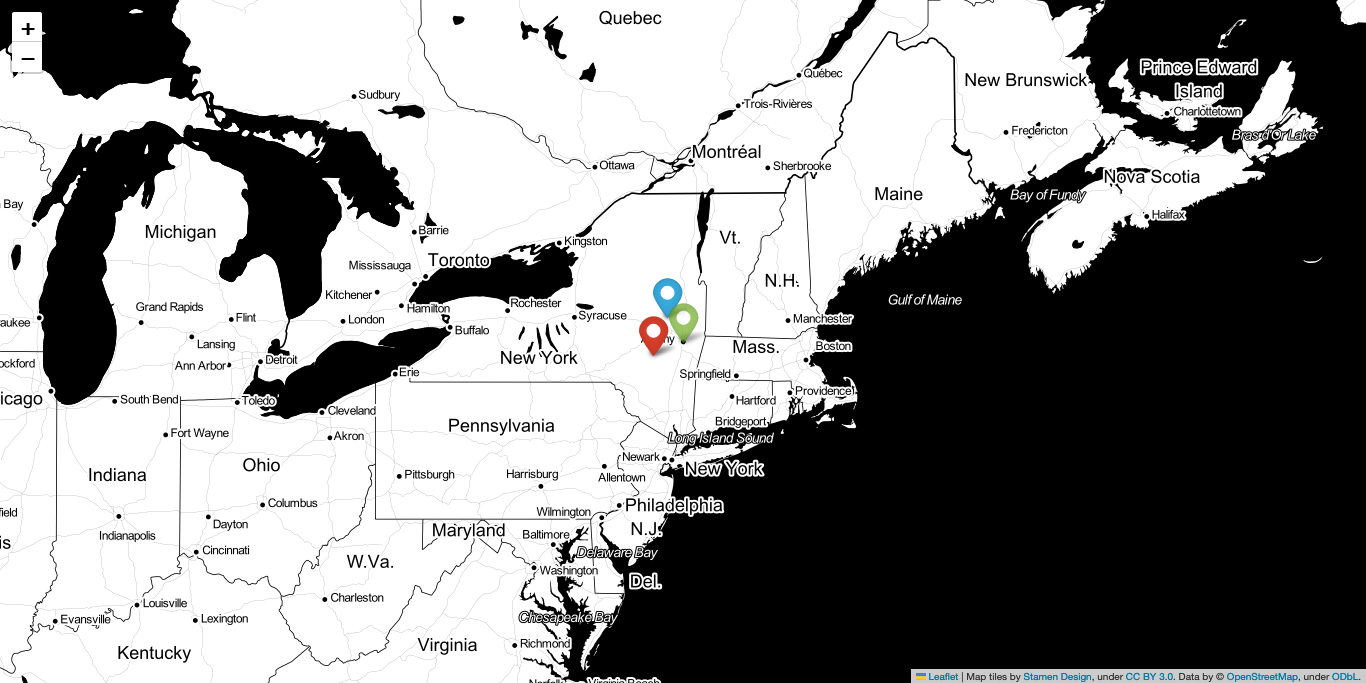}
         \caption{Albany, New York}
         \label{fig:map_coods_albany}
     \end{subfigure}
     \hspace{1cm}
    \begin{subfigure}[b]{0.435\textwidth}
         \centering
         \includegraphics[trim={10cm 6cm 10cm 5.5cm}, clip,  width=\textwidth]{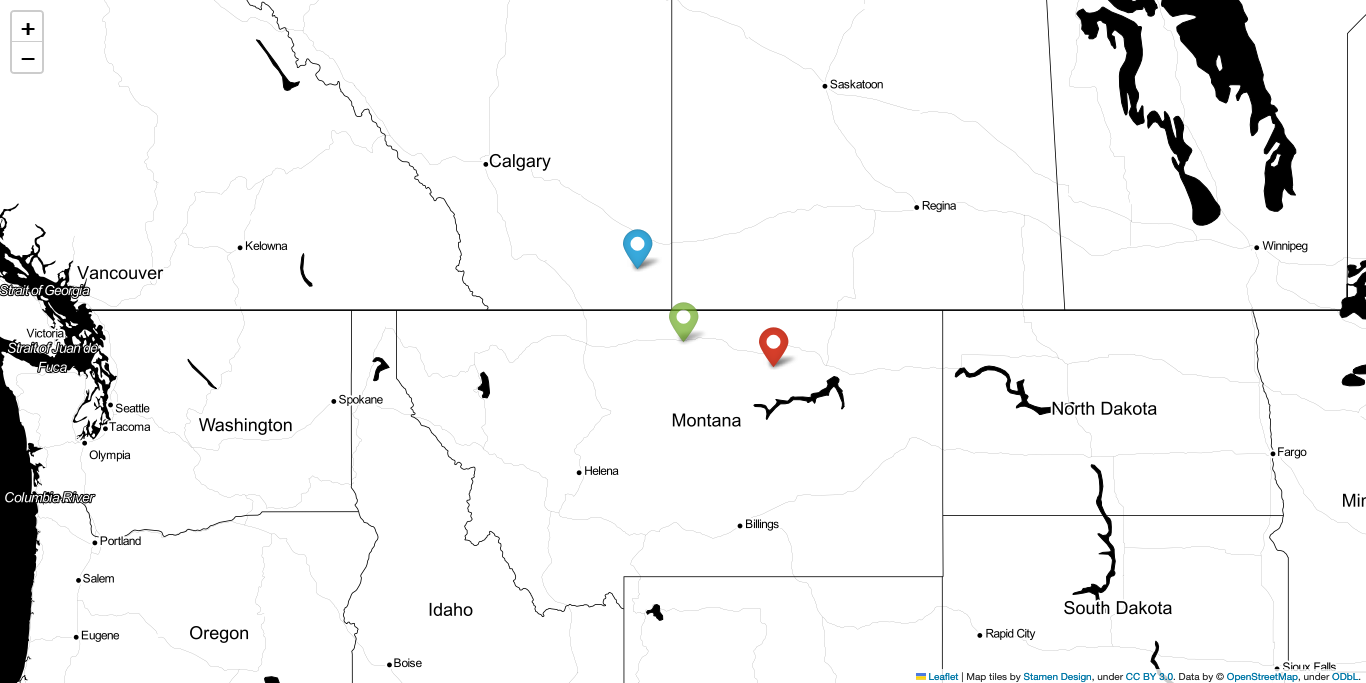}
         \caption{Havre, Montana}
         \label{fig:map_coods_havre}
     \end{subfigure}
     \begin{subfigure}[b]{0.435\textwidth}
         \centering
         \includegraphics[trim={10cm 6cm 10cm 5.5cm}, clip,  width=\textwidth]{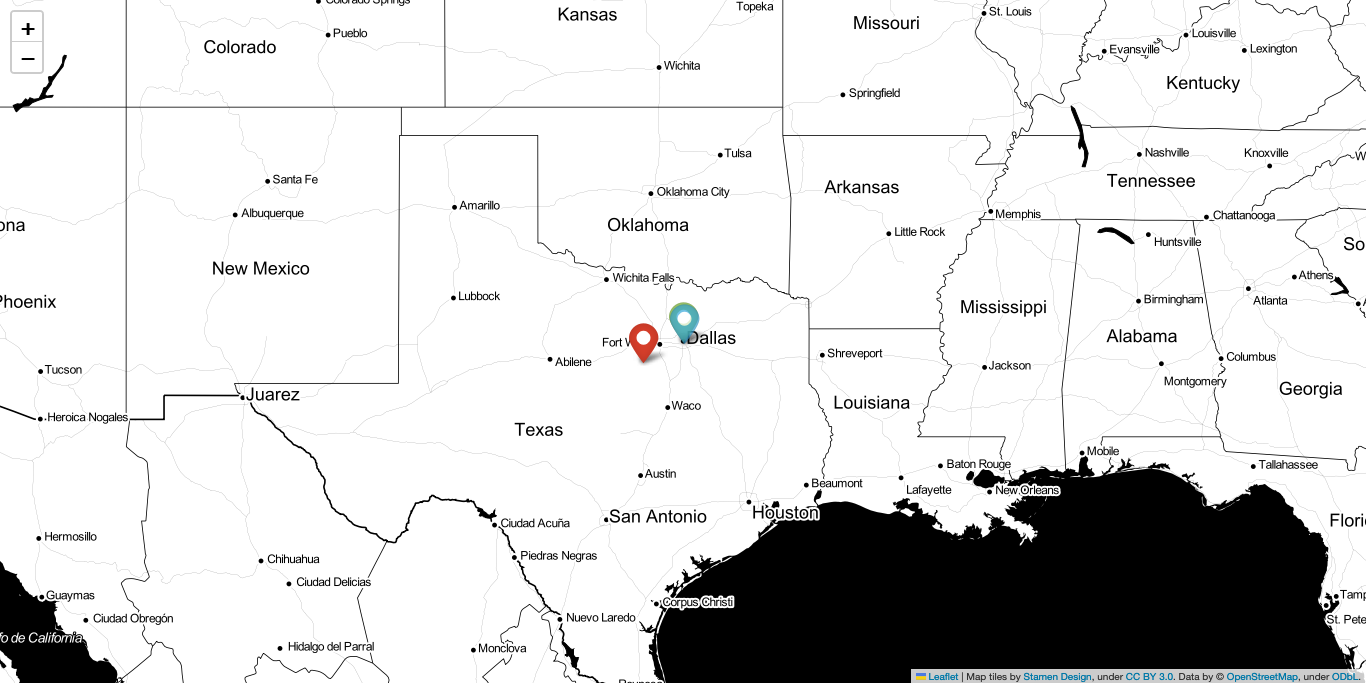}
         \caption{Dallas, Texas}
         \label{fig:map_coods_dallas}
     \end{subfigure}
     \hspace{1cm}
     \begin{subfigure}[b]{0.435\textwidth}
         \centering
         \includegraphics[trim={10cm 6cm 10cm 5.5cm}, clip,  width=\textwidth]{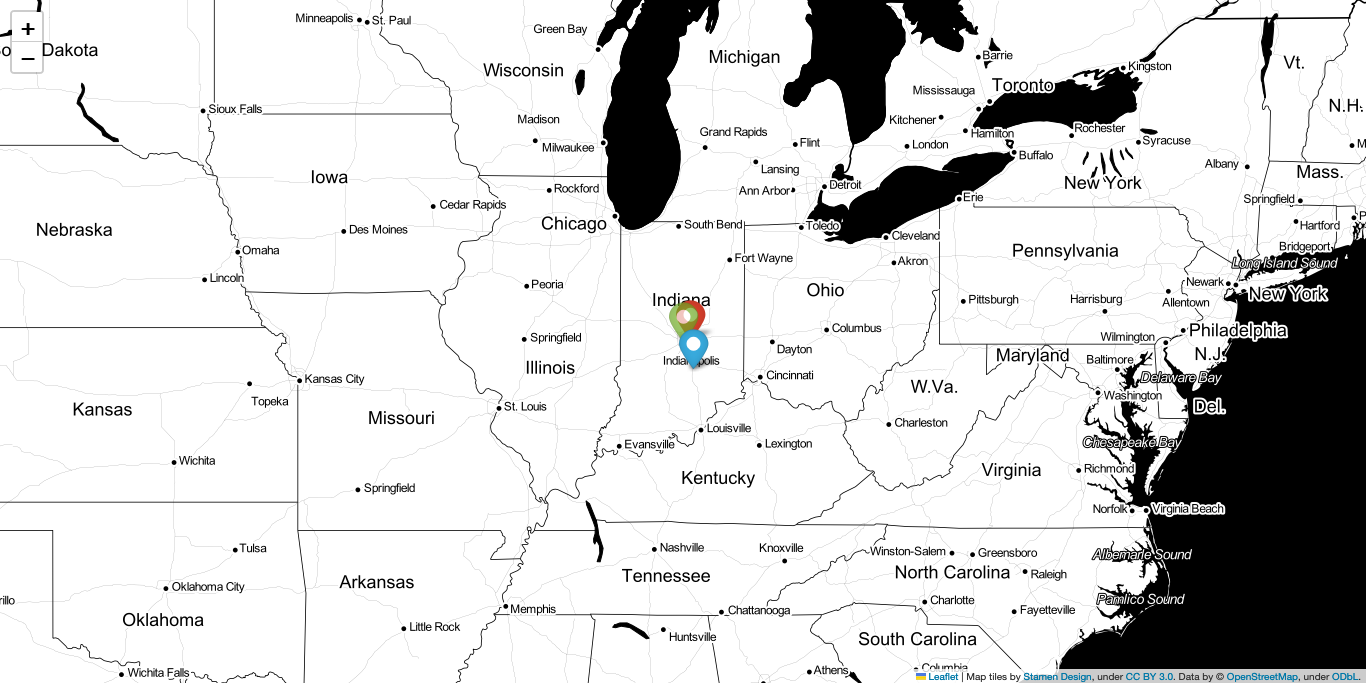}
         \caption{Indianapolis, Indiana}
         \label{fig:map_coods_indianapolis}
     \end{subfigure}
 
    \caption{Original and predicted locations based on actual distances and predicted distances for (a) Albany, New York, (b) Havre, Montana, (c) Dallas, Texas, and (d) Indianapolis, Indiana. Green: Actual, Blue: Predicted based on actual distance, and Red: Predicted based on predicted distance.}
  \Description{The original geo-coordinates along with the predicted ones based on real distance and predicted distance for (a) Albany, New York, (b) Havre, Montana, (c) Dallas, Texas, and (d) Indianapolis, Indiana. Green: Actual, Blue: Predicted based on actual distance, and Red: Predicted based on predicted distance.}
    \label{fig:coors_pred_map}
\end{figure*}
It is important to note that the goal of this task was not to assess the ability of LLMs to predict the geo-coordinates of the cities.\footnote{We already conducted a separate geo-coordinates prediction task for cities around the globe with promising results (\S4).}
Instead, our objective is to evaluate LLMs' geospatial reasoning capabilities.
One potential use case of our task can be to predict the relative orientation of a city instead of its exact location.
We do believe that LLMs have potential for such a use case.

Figure~\ref{fig:coors_pred_map} shows the actual locations along with the predicted locations using both actual distances and predicted distances for four cities.
On the maps, the green, blue, and red markers represent the actual locations, the predicted locations based on actual distances, and the predicted locations from predicted distances, respectively.
In the case of ``Albany, New York'' and ``Havre, Montana''  the predicted locations, whether based on actual or predicted distances, deviate only slightly from the actual values.
In the case of ``Dallas, Texas'', the actual locations align closely with predicted locations based on actual distances, but differ from the ones predicted using predicted distances. 
However, for ``Indianapolis, Indiana'' the predicted locations based on predicted distances align well with the actual values, while those based on the actual distances deviate.
These inconsistencies would only be marginally noticeable in the context of city scales if our main object was the orientation of the city rather than its precise location.
While we acknowledge that the predicted locations based on predicted distances differ from those based on actual distances, they still are reasonably close.
A high Pearson correlation coefficient of \textbf{0.92} between actual distances and predicted distances further showcases the potential of LLMs for such tasks.

In conclusion, our results demonstrate the potential use of LLMs for geospatial reasoning tasks.
While it is important to note that the values produced by LLMs may not precisely match the actual values, they still show a remarkable level of similarity and exhibit a high correlation.
Thus, LLMs have great potential for supporting humans in geospatial reasoning and analysis tasks with targeted fine-tuning tailored to a certain use case.

\section{Conclusions and Future Work}
This work demonstrates notable improvements in LLMs' ability to handle geospatial data, due not only to the increasing size of models, but also facilitated by novel techniques such as instruction tuning.
We show that LLMs encode geospatial knowledge, which can be leveraged for tasks that are simple, such as obtaining coordinates and locations for cities by probing those models, or more complex, such as a quantitative understanding of spatial prepositions. All this information can be extracted and utilized using the proper ``querying'' techniques such as prompting.
%
%
We demonstrate that LLMs show potential for geospatial reasoning tasks, but further enhancements are needed to meet the desired accuracy and performance levels. 
Overall, LLMs have come a long way and now exhibit geospatial awareness when generating text. 

Future research will focus on examining the practical applicability of LLMs in real-world applications involving geospatial data, as well as utilizing even larger models, and doing so for languages other than English.

Experiment Code: \href{https://github.com/prabin525/spatial-llm}{https://github.com/prabin525/spatial-llm}.

\begin{acks}
This work has been supported by the National Science Foundation Grant No. IIS-2127901.
Additionally, this work was supported by resources provided by the Office of Research Computing, George Mason University and by the National Science Foundation (Awards Number 1625039 and 2018631).
\end{acks}

\bibliographystyle{ACM-Reference-Format}
\bibliography{main}




\end{document}